\documentclass[lettersize,journal]{IEEEtran}
\usepackage{amsmath,amsfonts}
\usepackage{array}
\usepackage[caption=false,font=normalsize,labelfont=sf,textfont=sf]{subfig}
\usepackage{textcomp}
\usepackage{stfloats}
\usepackage{url}
\usepackage{verbatim}
\usepackage{graphicx}
\def\BibTeX{{\rm B\kern-.05em{\sc i\kern-.025em b}\kern-.08em
    T\kern-.1667em\lower.7ex\hbox{E}\kern-.125emX}}
\usepackage{balance}

\usepackage[ruled,linesnumbered]{algorithm2e}
\usepackage{enumitem}
\usepackage{cite}
\usepackage{amssymb}
\usepackage{xcolor}
\usepackage{tabularx}
\usepackage{multirow}   
\usepackage{bigstrut}   
\usepackage{siunitx}
\usepackage{graphicx} 
\usepackage{array} 
\usepackage{booktabs} 
\usepackage[colorlinks=true, linkcolor=blue]{hyperref}
\usepackage{algpseudocode}      
\usepackage{soul, color, xcolor}
\usepackage{placeins}

\begin{document}
\title{TTRD3: Texture Transfer Residual Denoising Dual Diffusion Model for Remote Sensing Image Super-Resolution}

\author{Yide Liu, Haijiang Sun, Xiaowen Zhang, Qiaoyuan Liu, Zhouchang Chen, and Chongzhuo Xiao 
}
\maketitle

\begin{abstract}
Remote Sensing Image Super-Resolution (RSISR) aims to reconstruct high-resolution (HR) remote sensing (RS) images from low-resolution (LR) RS images inputs by overcoming the physical limitations of imaging systems. As a specialized task in the RS domain, RSISR provides critical support for downstream fine-grained ground object interpretation. Traditional RSISR methods face three major technical shortcomings: First, the extremely complex ground object features in wide-area RS images make precise extraction of multi-scale target features challenging. Second, the inherent imaging constraints of RS images lead to insufficient prior information, resulting in poor semantic consistency between SR reconstructions and real scenes. Third, existing SR methods struggle to balance the inherent trade-off between geometric accuracy and visual quality. To address these challenges, this paper proposes Texture Transfer Residual Denoising Dual Diffusion Model (TTRD3), a diffusion model (DM)-based SR framework. Specifically, the Multi-scale Feature Aggregation Block (MFAB) is designed to extract spatially heterogeneous features with scale variations through parallel heterogeneous convolutional kernels. The Sparse Texture Transfer Guidance (STTG) is developed to mine multi-scale sparse texture features from HR reference RS images of similar scenes. The Residual Denoising Dual Diffusion Model (RDDM) framework is constructed to synergistically optimize geometric accuracy and visual quality by integrating the deterministic modeling of residual diffusion with the diverse generation of noise diffusion. Experiments on multi-source RS dat asets demonstrate that TTRD3 significantly outperforms existing state-of-the-art (SOTA) methods in both geometric accuracy and visual quality. Quantitative results show 1.43\% improvement in LPIPS and 3.67\% enhancement in FID compared to the best-performing SOTA baselines. Our code and model are available at \url{https://github.com/LED-666/TTRD3}. 

\end{abstract}

\begin{IEEEkeywords}
sparse texture transfer, residual denoising dual diffusion model, super-resolution, remote sensing.
\end{IEEEkeywords}

\section{Introduction}
\IEEEPARstart{R}{emote} sensing technology has matured significantly, but its application remains constrained by the limitations of image spatial resolution in critical domains such as environmental monitoring\cite{r3}, resource exploration\cite{r2}, change detection\cite{r5}, and national security\cite{r1}. SR technology compensates for the physical limitations of imaging systems through algorithmic enhancements, serving as a core solution to overcome spatial resolution bottlenecks. However, the intrinsic characteristics of RS images pose significant challenges. First, the complex interactions between ground objects (e.g., buildings and vegetation) make multi-scale feature extraction difficult\cite{r8}. Second, irreversible high-frequency information loss during RS imaging and the scarcity of multi-source paired reference data lead to insufficient prior information\cite{r7}. Third, existing SR methods struggle to balance geometric fidelity and visual authenticity, often producing over-smoothed textures or repetitive unrealistic artifacts\cite{r6}.

\begin{figure}[t!]  
    \centering  
    \includegraphics[width=0.49\textwidth]{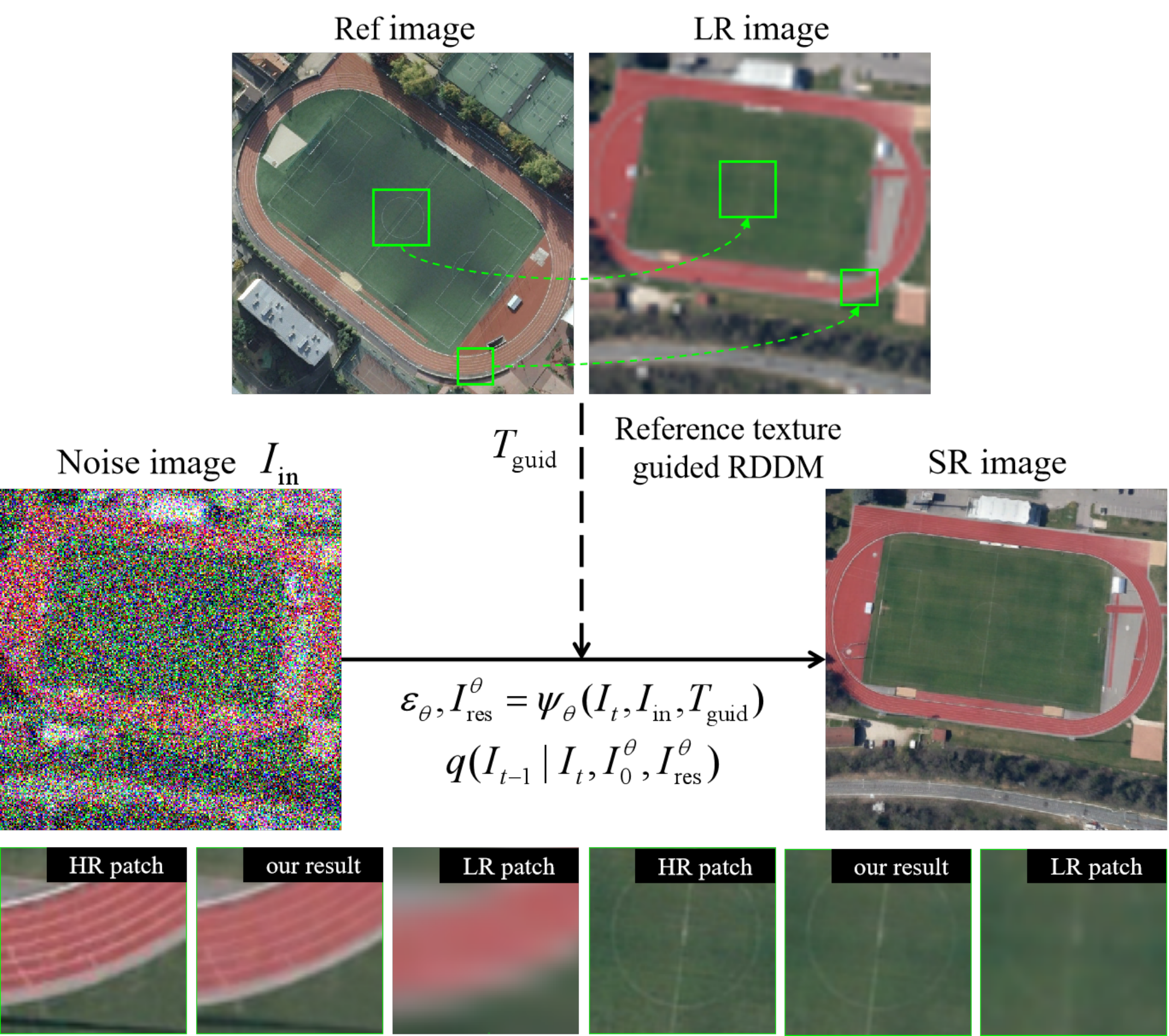}  
    \caption{The TTRD3 framework extracts structurally similar high-frequency spatial texture information from HR reference images, thereby guiding the RDDM to generate realistic high-frequency details.}  
    \label{figshow}  
\end{figure}

To address the challenge of feature extraction, current multi-scale feature extraction methods often fail to capture the hierarchical features of RS images. PMSRN\cite{PMSRN} employs dilated convolutions to expand receptive fields and uses a pyramidal multi-scale residual network to enhance contextual interactions, effectively addressing multi-scale ground object analysis. However, due to significant sampling intervals in convolutional kernels, it suffers from local information loss and limited performance in extracting small targets (e.g., roads, vehicles). WTCRR\cite{WTCRR} utilizes wavelet transforms in the frequency domain to extract periodic textures and suppress noise but lacks dynamic scale adaptation for multi-scale targets in complex scenes. ESTNet\cite{ESTNet} and TransENet\cite{TransENet} improve long-range dependency modeling through Transformer\cite{transformer} architectures but lack hierarchical design in multi-scale feature aggregation, leading to imbalanced feature weighting between small targets and large scenes.

Current prior-guided SR methods typically rely on auxiliary data to compensate for detail loss. For instance, DI-GAN\cite{DI-GAN} and UPSNet\cite{UPSNet} reconstruct HR spectral images using LR multispectral and panchromatic data. RGTGAN\cite{RGTGAN} leverages multi-view RS images as references, while Ref-Diff\cite{Ref-diff} utilizes multi-temporal RS image pairs and change detection masks as priors. However, these methods require strict spatial alignment between LR and reference images, which are often unavailable in real-world scenarios. Although CycleGAN-based\cite{CycleGAN} unsupervised cross-modal fusion relaxes alignment constraints, it may introduce semantic inconsistencies. To address this, we explore sparse texture transfer from unpaired HR reference images, injecting reasonable high-frequency components through similarity-guided diffusion without requiring pixel-level correspondence.

Existing SR frameworks face a dilemma: regression-based convolutional neural networks (CNNs)\cite{R11} and Transformer methods rely on deterministic approximations of LR-to-HR mappings, prioritizing geometric accuracy but producing blurred outputs\cite{r8}. Conversely, Generative adversarial networks (GANs\cite{r16})-based and diffusion model (DM\cite{r21})-based methods enhance visual quality at the cost of structural distortion. Specifically, GANs suffer from mode collapse, leading to repetitive textures\cite{GANreview}, while DMs lack explicit geometric constraints, causing deformation of ground objects. Balancing geometric precision and visual quality remains an urgent challenge.

To address the aforementioned challenges, this study proposes the Texture Transfer Residual Denoising Dual Diffusion Model (TTRD3), establishing a novel solution framework for RSISR tasks as illustrated in Figure \ref{figshow}. Specifically, to extract multi-scale information from both reference and LR images, we design a Multi-scale Feature Aggregation Block (MFAB) employing parallel heterogeneous convolutional kernels, integrated with the Convolutional Block Attention Module (CBAM) to dynamically weight critical scale features, thereby significantly enhancing the feature representation capability for complex ground objects. To compensate for the inherent high-frequency information loss in remote sensing LR images, we additionally introduce HR reference images from similar scenes. The Sparse Texture Transfer Guidance (STTG) is constructed to extract strongly correlated sparse texture features from HR reference images through a soft-hard attention collaborative transfer mechanism. Here, soft attention models fine-grained texture similarity via local-global hybrid correlation matrices, while hard attention suppresses redundant noise through dynamic Top-K sparse selection to improve reconstruction accuracy in key regions. To resolve the challenging trade-off between geometric precision and visual quality, we develop the Residual Denoising Dual Diffusion Super-Resolution (RDDM) architecture. RDDM effectively balances the accuracy-authenticity dilemma in DM-based remote sensing SR tasks by combining the deterministic modeling of residual diffusion with the diverse generation capability of noise diffusion. The primary contributions of this study are summarized as follows:
\begin{enumerate}[label=\arabic*)]
    \item Design an MFAB feature extraction module to enhance multi-scale feature representation by integrating heterogeneous convolution and the CBAM attention mechanism.
    \item Propose an STTG texture migration method that replaces spatial alignment with deep semantic similarity, adaptively selecting strongly correlated sparse texture information to address the challenge of insufficient prior information in SR tasks.
    \item Develop an RDDM-based SR framework that leverages deterministic residual diffusion to constrain the diversity of noise diffusion, achieving a balance between geometric accuracy and visual quality.
    \item The proposed TTRD3 is proposed, and evaluations on the AID and RSD46 public datasets demonstrate that TTRD3 surpasses previous SOTA results with only 10 inference steps.
\end{enumerate}

The remainder of this paper is organized as follows. Section \ref{sec2} provides an overview of the advancements in SR tasks. Section \ref{sec3} elaborates on the technical details and highlights the innovations of our proposed TTRD3 framework. Section \ref{sec4} describes the datasets utilized and the experimental details. Section \ref{sec5} presents the experiments and analysis. Finally, Section \ref{sec6} concludes the paper with a summary of the key contributions and outlines potential future research directions.

\section{Related Work}\label{sec2}
\subsection{SR models based on regression}
With the advent of the deep learning revolution, regression-based deep SR methods have gradually dominated the field. SRCNN\cite{SRCNN} first constructed a three-layer convolutional network, breaking the performance bottleneck of traditional algorithms, but its shallow structure struggled to capture deep-level features. VDSR\cite{VDSR} overcame network depth limitations through residual learning, enabling deeper SR networks. The introduction of the ResNet\cite{RESNET} architecture gave rise to deep residual networks like SRResNet\cite{SRGAN}, achieving higher performance metrics. SRDenseNet\cite{SRDenseNet} adopted dense connections to enhance feature reuse. DRCN\cite{DRCN} and DRRN\cite{DRRN} employed recursive structures to increase network depth while reducing computational complexity and storage costs through parameter sharing. CARN\cite{CARN} built a lightweight recursive model for more efficient modeling. Although CNN-based methods have been widely implemented and achieved outstanding performance across tasks, their equal treatment of channels via convolutional kernels limits the utilization of high-frequency information. Attention mechanisms emerged as a solution, enabling deeper interdependencies among channels. RCAN\cite{RCAN} introduced channel attention modules that assign varying weights to each channel by learning global information from input feature maps. SAN\cite{SAN} proposed a channel attention mechanism based on second-order statistics, enhancing feature correlations for more discriminative representations. The introduction of Transformer architectures brought breakthroughs in global modeling: SwinIR\cite{Swinir} achieved efficient long-range dependency capture through hierarchical window self-attention. TTSR\cite{TTSR} pioneered reference image-guided paradigms to mitigate high-frequency information loss. ESRT\cite{ESRT} balanced performance and computational costs via a CNN-Transformer hybrid architecture, and ELAN\cite{ELAN} optimized attention-sharing mechanisms to improve parameter efficiency. In RSISR tasks, ERCNN\cite{ERCNN} proposed a dual-brightness scheme to optimize feature retention and cross-feature map discriminability. FGRDN\cite{FGRDN} employed novel feedback mechanisms and ghost modules for low-parameter feature refinement. MAST\cite{MAST} designed multi-scale attention to enhance geographical feature perception, while TTST\cite{TTST} established a token selection mechanism to dynamically choose top-k critical tokens for more effective and compact long-range modeling. However, these regression methods centered on L1 (Mean Absolute Error) and L2 (Mean Square Error) loss functions, while optimizing image reconstruction accuracy through pixel-level difference minimization, inherently force generated results to converge toward "average solutions" of target images, leading to high-frequency texture loss and edge blurring—particularly evident in complex scenes like foliage and architectural textures\cite{RRR}.

\subsection{SR models based on GANs} 
In recent years, GANs have achieved remarkable progress in SR, opening new technical pathways to break through the limitations of traditional regression methods. SRGAN\cite{SRGAN} first introduced adversarial training to SR, synthesizing highly realistic textures through generator-discriminator game mechanisms, significantly enhancing visual quality. ESRGAN\cite{ESRGAN} improved upon SRGAN by designing Residual-in-Residual Dense Blocks (RRDB) and removing batch normalization layers, while introducing perceptual loss functions that leverage pretrained VGG models to align generated images with real ones in feature space. Real-ESRGAN\cite{real-ESRGAN} extended ESRGAN to handle real-world degradation patterns. SAGAN\cite{SAGAN} integrated cross-image attention mechanisms to establish long-range, multi-level dependencies for detail generation. SRNTT\cite{SRNTT} proposed a reference-guided patch-matching strategy that utilizes multi-level patch matching to incorporate reference image information during SR. LDL\cite{LDL} enhanced image authenticity through local discriminative learning to distinguish GAN-generated artifacts from real details. For RSISR tasks, SCSE-GAN\cite{SCSE-GAN} stacked Spatial and Channel Squeeze-Excitation (SCSE) blocks after residual layers and adopted Wasserstein GAN with gradient penalty (WGAN-GP) to stabilize training. MA-GAN\cite{MA-GAN} developed a multi-attention framework with attention-based fusion blocks for improved geographical feature representation. SRAGAN\cite{SRAGAN} optimized adversarial learning by capturing structural components and long-term dependencies through local and global attention mechanisms. TE-SAGAN\cite{TE-SAGAN} introduced self-attention modules and texture loss units to refine edges and textures. RGTGAN\cite{RGTGAN} proposed gradient-assisted GANs to enhance textures via gradient branches. Although GAN-based methods excel in perceptual metrics like LPIPS and hybrid loss strategies have substantially improved texture generation, challenges remain in computational complexity and training stability. DMs, with their progressive generation paradigm, offer new solutions by achieving high-quality reconstruction without mode collapse, emerging as a promising direction for next-generation SR technology.

\subsection{SR models based on diffusion models} 
DMs establish a rigorous mathematical framework for image generation through dual processes: forward noise perturbation and reverse denoising learning. DDPM\cite{DDPM}, as a milestone work, first proposed a trainable Markov chain diffusion architecture. DDIM\cite{DDIM} introduced non-Markovian accelerated sampling strategies to significantly improve inference efficiency. RDDM\cite{RDDM} changes the traditional diffusion paradigm by additionally introducing residual information to constrain the generation results, achieving remarkable effects in the fields of generation, image dehazing, and denoising. Classifier guidance\cite{classifier} and classifier-free guidance\cite{classifier-free} techniques achieved comprehensive quality superiority over GANs through conditional control mechanisms. Stable Diffusion\cite{SD} addresses the computational costs and accuracy constraints of previous approaches by developing a Latent Diffusion Model, which mitigates the challenges of building diffusion models directly on high-dimensional feature spaces. Compared with traditional GANs and VAEs, DMs demonstrate significant advantages in modeling complex image statistical distributions, with their progressive generation characteristics effectively avoiding mode collapse.

SR3\cite{SR3} pioneered the application of DMs to SR tasks, achieving HR image reconstruction through end-to-end noise prediction. SRDiff\cite{SRDiff} proposed a residual diffusion paradigm by predicting LR-HR residual mappings, and ResShift\cite{Resshift} subsequently built an efficient residual chain generation architecture. For RS images characteristics, TESR\cite{TESR} developed a two-stage enhancement SR framework, first validating the feasibility of DMs in RSISR. EDiffSR\cite{Ediffsr} innovatively designed a Conditional Prior Enhancement Module (CPEM) and Efficient Activation Network (EANet), substantially reducing computational complexity. FastDiffSR\cite{FastDiffSRl} constructed a conditional diffusion acceleration framework, achieving breakthrough improvements in inference speed.

However, most existing DM-based RSISR methods focus primarily on designing new conditional guidance modules and improving noise predictors, while neglecting the introduction of additional high-frequency information and refinement of diffusion frameworks. This paper proposes an optimized residual noise double diffusion method, which effectively balances the inherent contradiction between geometric accuracy and visual quality, and innovatively designs the self - adaptive texture migration mechanism STTG. The method supplements RDDM with additional high-frequency prior guidance information by transferring strongly correlated texture features from reference images.

\section{Method}\label{sec3}
\subsection{Residual Denoising Dual Diffusion Model} 

\begin{figure*}[t!]  
    \centering
    \includegraphics[width=1\textwidth]{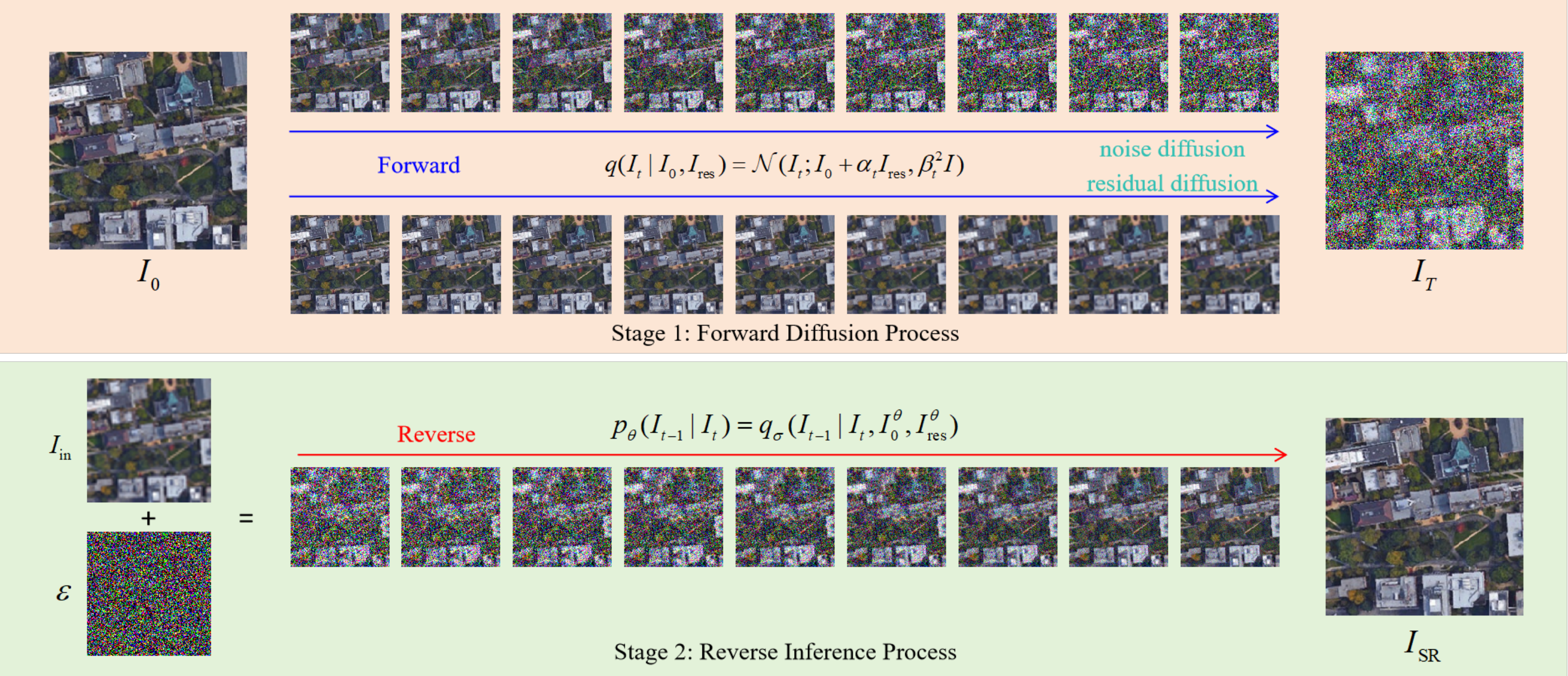}  
    \caption{illustrates the forward process and reverse inference process of the RDDM, comprising two components: residual diffusion and noise diffusion. Here, $I_0$ represents the HR image, $I_T$ denotes the dual-diffusion degraded image, $I_{\text{in}}$ corresponds to the residual degraded image (LR image), $I_{\text{SR}}$ denotes the SR image, and $\varepsilon$ signifies the gaussian noise.}
    \label{figdiff}
\end{figure*}

To address the special requirements of reconstruction fidelity in RSISR tasks, the RDDM simultaneously diffuses residual information and noise information during the forward diffusion process. Specifically, residual diffusion emphasizes determinism by constraining the spatial fidelity of the reconstruction process through directional propagation of residual information. Noise diffusion focuses on diversity, representing stochastic perturbations in the diffusion process to ensure reconstruction diversity. The forward diffusion process and reverse inference process of RDDM will be elaborated in detail below.
\subsubsection{Forward diffusion process}
The forward process of RDDM allows for the simultaneous diffusion of residuals and noise, with the target image gradually diffusing into a degraded version carrying noise, where a single forward diffusion step is defined as:
\begin{equation}\tag{1}
    I_t = I_{t-1} + I_{\text{deg}}^t,\quad I_{\text{deg}}^t\sim \mathcal{N}(\alpha_tI_{\text{res}},\beta_t^2I)\label{eq1}
\end{equation}

Here, \( I_{\text{deg}}^t \) represents the degradation information from state \( I_{t-1} \) to state \( I_t \), which includes residual diffusion and noise diffusion. \( I_{\text{res}} \) is the difference between the input image \( I_{\text{in}} \) and the original image \( I_0 \). 
\begin{equation}\tag{2}
    I_{\text{res}} = I_{\text{in}}-I_0
\end{equation}

Unlike DDPM, which uses a single coefficient schedule to control the mixing ratio of noise and images, RDDM employs two independent coefficient schedules, \(\alpha_t\) and \(\beta_t\) control the diffusion rates of residuals and noise, respectively. The specific method is as follows:
\begin{align*}
    I_t &= I_{t - 1} + \alpha_t I_{\text{res}} + \beta_t \varepsilon_{t - 1}\\
    &= I_{t - 2} + (\alpha_{t - 1} + \alpha_t)I_{\text{res}} + (\sqrt{\beta_{t - 1}^2 + \beta_t^2})\varepsilon_{t - 2}\\
    &=\cdots\\
    &=I_0 + \bar{\alpha}_t I_{\text{res}} + \bar{\beta}_t \varepsilon\tag{3}\label{eq3}
\end{align*}

Here, $\varepsilon_t,\varepsilon_{t-1},\cdots,\varepsilon\sim \mathcal{N}(0,I);\bar{\alpha}_t = \sum_{i=1}^t \alpha_t;
\bar{\beta}_t = \sqrt{\sum_{i=1}^t \beta_t^2}$. if $t = T$, then $\bar{\alpha}_T=1$ and $I_T = I_{\text{in}} + \bar{\beta}_T \varepsilon$, Among them, $\bar{\beta}_T$ is a manually set parameter used to control the intensity of noise.

\begin{figure*}[t!]  
    \centering
    \includegraphics[width=\textwidth]{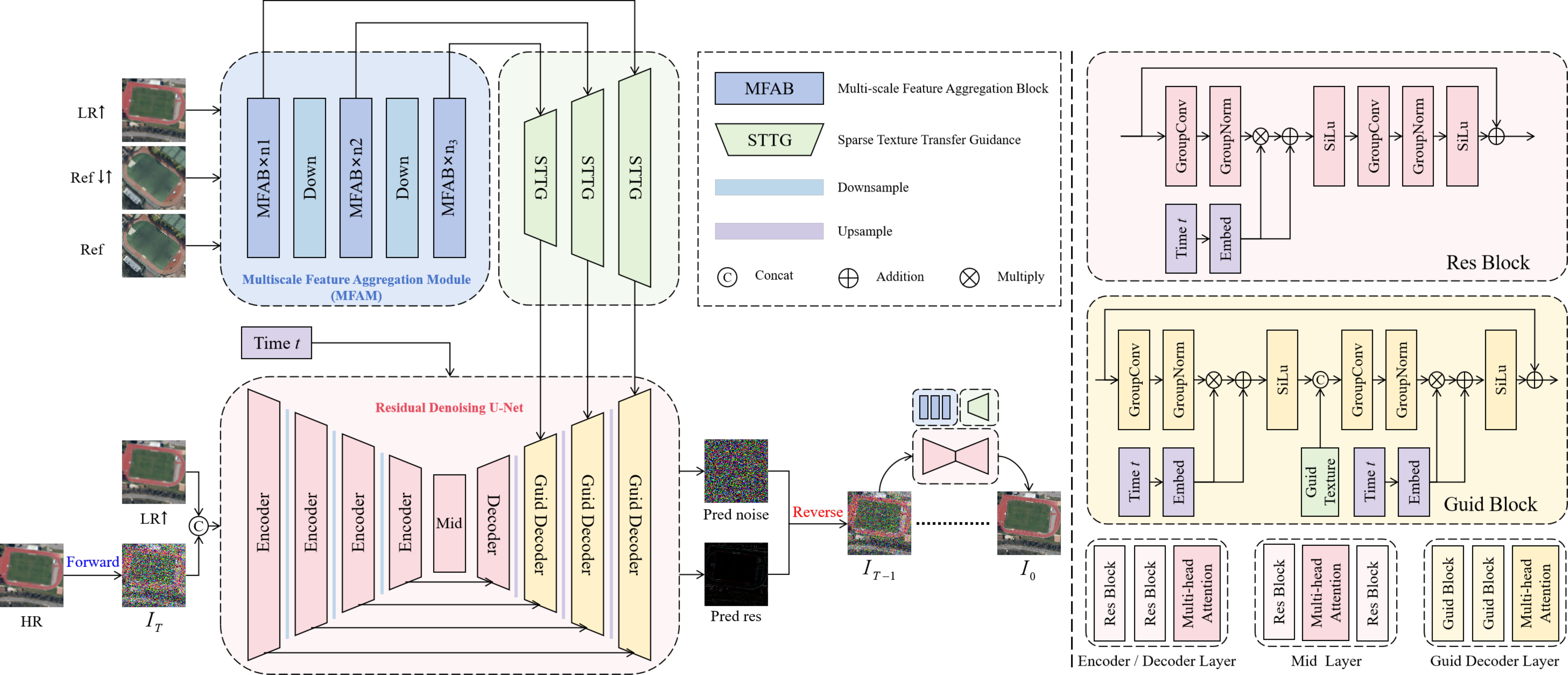}  
    \caption{The overall framework of the proposed TTRD3 is illustrated. Here, LR↑ denotes the 4× upsampled LR image, Ref↓↑ represents the degraded reference image obtained by downsampling and subsequent upsampling, and Ref is the high-quality reference image. These inputs are processed by the MFAM to extract multi-scale features at three levels. The STTG then generates multi-scale sparse texture guidance maps from these features. Finally, the Guid Decoder integrates the texture guidance information into the residual denoising U-Net network for high-fidelity reconstruction.}
    \label{fig3}
\end{figure*}

From equation \ref{eq1}, it can be seen that the joint probability distribution of the forward diffusion process is defined as:
\begin{equation}\tag{4}
    q(I_{1:T}\vert I_0,I_{\text{res}}) =\prod_{t = 1}^{T}q(I_t\vert I_{t - 1},I_{\text{res}})
\end{equation}
\begin{equation}\tag{5}
    q(I_t\vert I_{t - 1},I_{\text{res}}) =\mathcal{N}(I_t;I_{t - 1}+\alpha_tI_{\text{res}},\beta_t^2I)
\end{equation}

Equation \ref{eq3} defines the marginal probability distribution as:
\begin{equation}\tag{6}
    q(I_t\vert I_0, I_{\text{res}}) =\mathcal{N}(I_t;I_0+\alpha_t I_{\text{res}},\beta_t^2 I)
\end{equation}

\subsubsection{Reverse inference process}

In the forward process of equation \ref{eq3}, the residual ($I_{\text{res}}$) and noise ($\varepsilon$) are gradually added to $I_0$ to form $I_t$, while in the inverse process from $I_T$ to $I_0$, the $I_{\text{res}}$ and $\varepsilon$ need to be estimated from the network for subtraction. The $I_{\text{res}}$ and $\varepsilon$ can be estimated by the network. Training networks and using the input image $I_t$, $I_t$ can be synthesized using $I_0$, $I_{\text{res}}$ and $\varepsilon$ by equation \ref{eq3}. 

The target image $I_0^\theta$ can be estimated according to equation \ref{eq3}.
\begin{equation}\tag{7}
    I_0^\theta = I_t - \bar{\alpha}_t I_{\text{res}}^\theta - \bar{\beta}_t \varepsilon^\theta
\end{equation}

If $I_0^\theta$ and $I_{\text{res}}^\theta$ are given, the reverse inference process is defined as: 
\begin{equation}\tag{8}\label{eq8}
    p_\theta(I_{t-1} \vert I_t) = q_\sigma(I_{t-1} \vert I_t, I_0^\theta, I_{\text{res}}^\theta)
\end{equation}

where the transition probability from $I_t$ to $I_{t-1}$ is:
\begin{align*}
    q_{\sigma}(I_{t-1}&|I_{t},I_{0},I_{\text{res}})
    =\mathcal{N}(I_{t-1};I_{0}+\bar{\alpha}_{t-1}I_{\text{res}}\\+
    &\quad\sqrt{\bar{\beta}_{t-1}^{2}-\sigma_{t}^{2}} \frac{I_{t}-(I_{0}+\bar{\alpha}_{t-1}I_{\text{res}})}{\bar{\beta}_{t}},\sigma_{t}^{2}I)\tag{9}\label{eq9}
\end{align*}

where $\sigma_{t}^{2}=\eta\beta_{t}^{2}\bar{\beta}_{t-1}^{2}/\bar{\beta}_{t}^{2}$, and $\eta$ controls the randomness ($\eta = 1$) or determinacy ($\eta = 0$) of the generation process.

From equation \ref{eq3} and equation \ref{eq8}, it can be seen that equation \ref{eq9} can be written as:
\begin{align*}
    I_{t - 1} &= I_0+\bar{\alpha}_{t - 1}I_{\text{res}}^\theta+\\[4pt]
    &\quad\quad\sqrt{\bar{\beta}_{t - 1}^2-\sigma_t^2}\frac{I_t-(I_0+\bar{\alpha}_tI_{\text{res}}^\theta)}{\bar{\beta}_t}+\sigma_t\varepsilon_t\\
    &=\frac{\sqrt{\bar{\beta}_{t - 1}^2-\sigma_t^2}}{\bar{\beta}_t}I_t+(1-\frac{\sqrt{\bar{\beta}_{t - 1}^2-\sigma_t^2}}{\bar{\beta}_t})I_0+\\
    &\quad\quad(\bar{\alpha}_{t - 1}-\frac{\sqrt{\bar{\alpha}_t\bar{\beta}_{t - 1}^2-\sigma_t^2}}{\bar{\beta}_t})I_{\text{res}}^\theta+\sigma_t\varepsilon_t\\
    &=I_t\!\!-\!(\bar{\alpha}_t\!\!-\!\bar{\alpha}_{t - 1})I_{\text{res}}^\theta\!\!-\!(\bar{\beta}_t\!\!-\!\!\sqrt{\bar{\beta}_{t \!-\! 1}^2\!-\!\sigma_t^2})\varepsilon_\theta\!\!+\!\!\sigma_t\varepsilon_t\tag{10}\label{eq10}
\end{align*}

Where $\varepsilon_t\sim \mathcal{N}(0,I)$. Equation \ref{eq10} is the sampling formula for the reverse process.

Similar to the evolution from DDPM to DDIM, an inductive argument from time $T$ to time $1$ can be used here to prove that the following equation holds at any time $t$.
\begin{equation}\tag{11}
    q(I_t\vert I_0,I_{\text{res}}) = \mathcal{N}(I_t;I_0+\bar{\alpha}_tI_{\text{res}},\bar{\beta}_t^2I)
\end{equation}

Therefore, the final reverse sampling formula is:
\begin{equation}\tag{12}\label{eq12}
    I_{\text{prev}}\!=\!I_t\!-\!(\bar{\alpha}_t\!-\!\bar{\alpha}_{\text{prev}})I_{\text{res}}^\theta\!-\!(\bar{\beta}_t\!-\!\sqrt{\bar{\beta}_{\text{prev}}^2\!-\!\sigma_t^2})\varepsilon_\theta\!+\!\sigma_t\varepsilon
\end{equation}

Among them, there can be multiple iteration steps between $I_{\text{prev}}$ and $I_t$.

A schematic diagram of the forward process and reverse inference of the RDDM is shown in Figure \ref{figdiff}.

\subsection{Overview} 
The workflow of the TTRD3 network framework is illustrated in Figure \ref{fig3}. The network inputs include the bicubic-interpolated LR remote sensing image LR↑, the original HR reference image Ref, and its degraded version Ref↓↑. These inputs are first processed by the Multi-scale Feature Aggregation Module (MFAM), which employs a pyramid architecture to hierarchically extract multi-scale ground object features (e.g., buildings, vegetation) through parallel multi-scale convolutional kernels in the MFAB. This hierarchical extraction ultimately generates three distinct scales of deep feature maps.

Subsequently, the upsampled LR↑ image and the multi-scale features from the reference image group are fed into the STTG module. This module constructs local-global similarity matrices between LR and Ref feature patches using a soft-hard attention collaborative mechanism, combined with a dynamic Top-K selection strategy to filter strongly correlated semantic textures. This process generates multi-scale sparse texture guidance maps.

In the RDDM stage, the texture guidance information is injected into the residual denoising U-Net network via the Guid Decoder layer. During the diffusion process, the LR image and the residual noise-degraded image $I_T$ are used as conditional inputs, jointly with the texture guidance information, to predict the residual and noise components. Finally, the high-fidelity SR image is reconstructed through an iterative reverse diffusion process that removes noise and fuses residual information.

Notably, to better balance performance and computational efficiency, we designed three network variants with varying capacities (TTRD3-1Net-A/B and TTRD3-2Net). Detailed performance comparisons and efficiency analyses will be comprehensively discussed in Section \ref{sec5}.

\subsection{Multiscale Feature Aggregation Block} 

\begin{figure}[t!]  

    \centering  
    \includegraphics[width=0.33\textwidth]{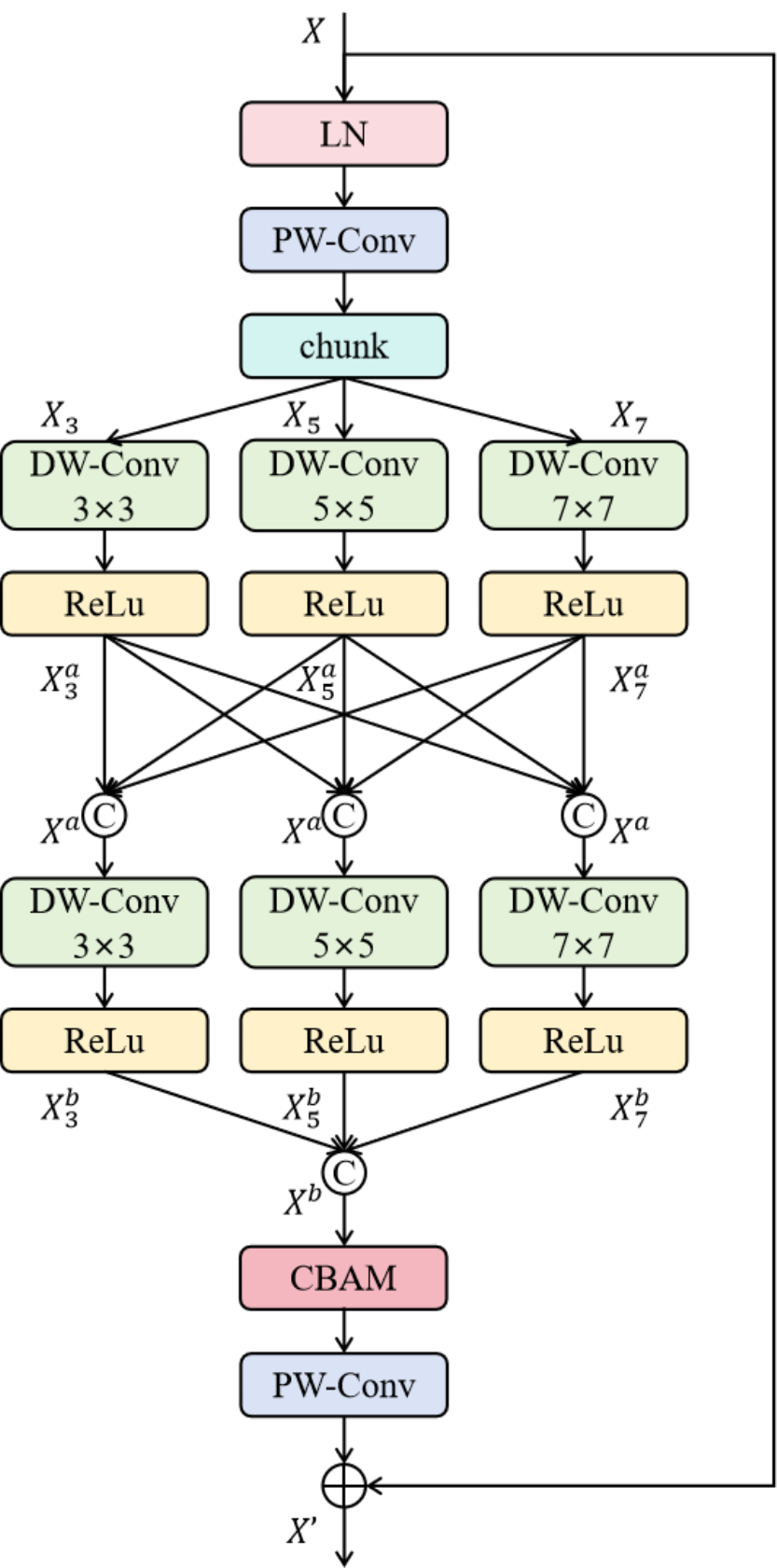}  
    \caption{illustrates the Multi-scale Feature Aggregation Block(MFAB)}  
    \label{figMFAB}  
\end{figure}

\begin{figure}[t!]  
    \centering  
    \includegraphics[width=0.48\textwidth]{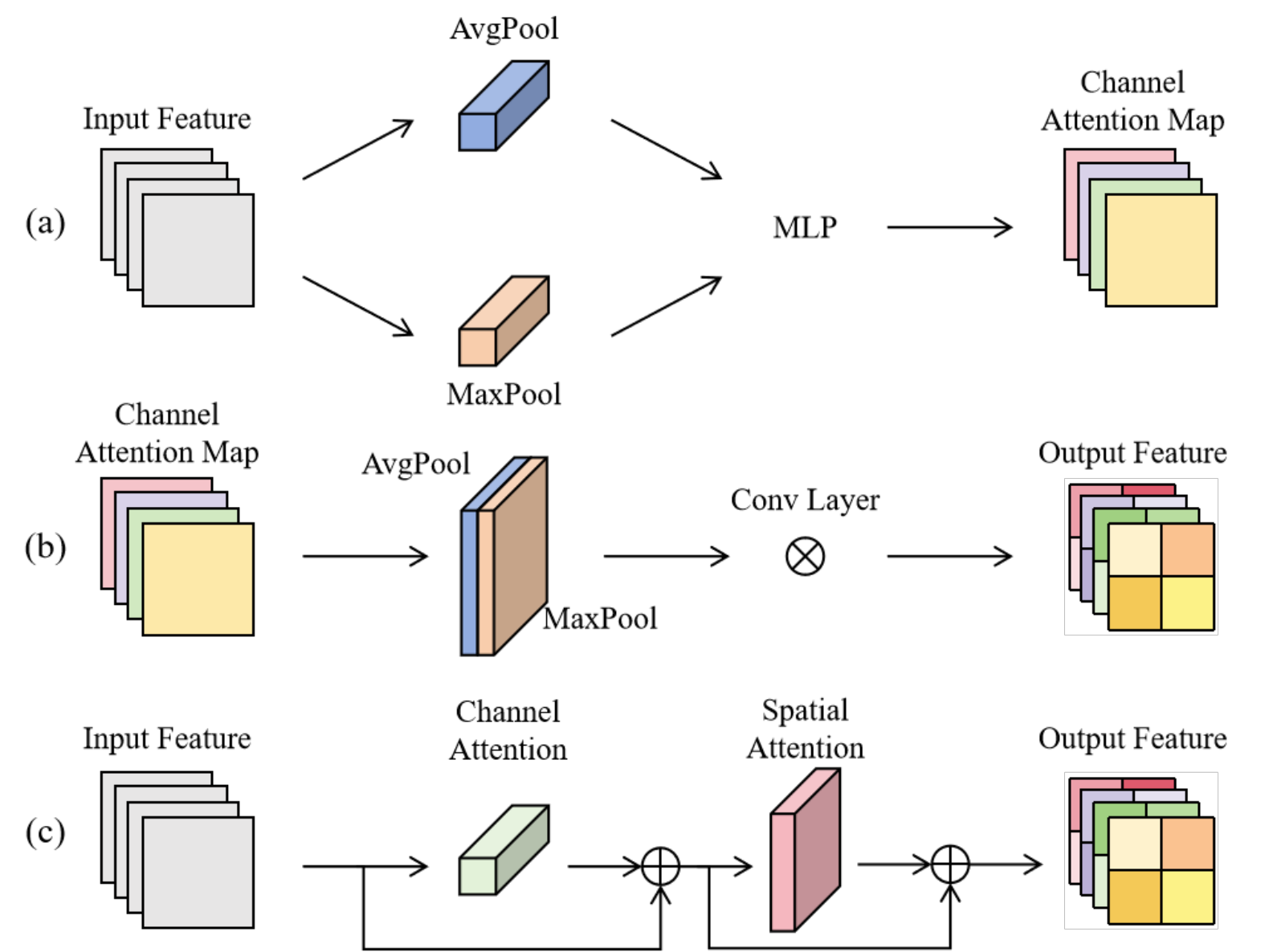}
    \caption{The structural diagram of CBAM. (a) CAM, (b)SAM, (c) overall process of CBAM.}  
    \label{figCBAM}  
\end{figure}

To address multi-scale inconsistency between reference and LR images, we design the MFAB with the detailed structure shown in Figure \ref{figMFAB}. The block captures features across spatial hierarchies through parallel multi-scale convolutional kernels and integrates the CBAM (shown in Figure \ref{figCBAM}) to dynamically fuse spatial and channel attention mechanisms, thereby addressing limitations in cross-scale feature fusion and noise suppression.

Specifically, given an input feature tensor $X\in\mathbb{R}^{H\times W \times C}$, the process begins with layer normalization (LN) followed by a point-wise (PW) convolution to adjust the channel dimension to $3\times C'$.  The tensor is then split along the channel dimension into three sub-tensors  ($X_3$, $X_5$, $X_7$), each with $C'$ channels.
\begin{equation}\tag{13}
    X_3,X_5,X_7=\text{chunk}(f_{1\times1}(\text{LN}(X)))
\end{equation}
Each sub-tensor is processed by a $3\times3$, $5\times5$, or $7\times7$ depth-wise (DW) convolution to extract multi-scale local features. The processed features are then concatenated to generate the shallow fused features $X^a$.
\begin{equation}\tag{14}
\begin{split}
\begin{cases}
X_{3}^{a} = \text{ReLU}(f_{3\times3}(X_{3})) \\
X_{5}^{a} = \text{ReLU}(f_{5\times5}(X_{5})) \\
X_{7}^{a} = \text{ReLU}(f_{7\times7}(X_{7})) 
\end{cases}\\
X^{a} = \text{Concat}(X_{3}^{a}, X_{5}^{a}, X_{7}^{a})
\end{split}
\end{equation}
Further enhanced through secondary multi-scale DW convolutions, the deep multi-scale features ($X_3^b$, $X_5^b$, $X_7^b$) are aggregated into $X^b$. $X^b$ is adaptively weighted by CBAM,  which adjusts spatial and channel-wise importance across different scales. Finally, a PW convolution restores the original channel dimension $C$, and a residual connection preserves the input's original information to generate the output $X'$.

\begin{equation}\tag{15}
\begin{split}
&\begin{cases}
X_{3}^{b} = \text{ReLU}(f_{3\times3}(X^a)) \\
X_{5}^{b} = \text{ReLU}(f_{5\times5}(X^a)) \\
X_{7}^{b} = \text{ReLU}(f_{7\times7}(X^a)) 
\end{cases}\\
&X^{b} = \text{Concat}(X_{3}^{b}, X_{5}^{b}, X_{7}^{b})\\
&X'= X + f_{1\times1}(f_{\text{CBAM}}(X^b))
\end{split}
\end{equation}

\subsection{Sparse Texture Transfer Guidance} 

\begin{figure*}[t!]  
    \centering
    \includegraphics[width=1\textwidth]{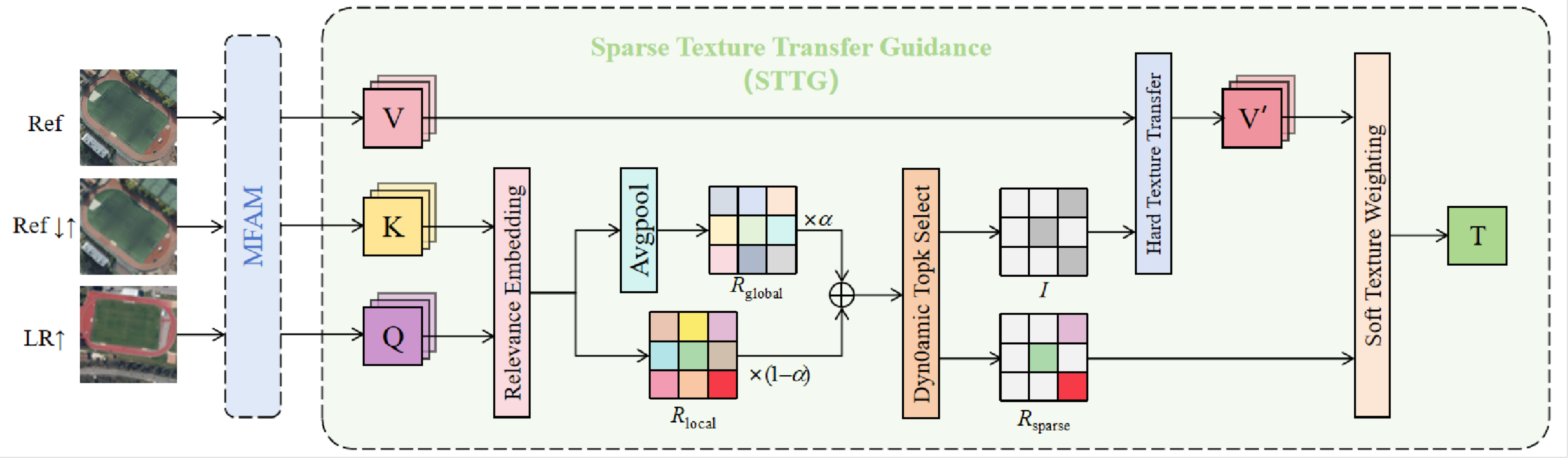}
    \caption{The STTG framework captures local details of texture while maintaining global perceptual consistency through a hybrid local-global soft attention mechanism. It dynamically balances the weights of global attention using a learnable parameter $\alpha$ and retains sparse correlation matrices and corresponding indices through a learnable Top-K hard attention mechanism. By combining soft and hard attention via matrix multiplication, it achieves adaptive texture transfer. }
    \label{figSTTG}
\end{figure*}

This study proposes a STTG method, its core innovation lies in constructing a collaborative framework combining a local-global hybrid soft attention mechanism and a learnable Top-K sparsification hard attention mechanism, as shown in Figure \ref{figSTTG}, enabling end-to-end cross-modal texture matching and transfer.

\subsubsection{Multi-scale Semantic Embedding}
To eliminate scale discrepancies between the reference image \( \text{Ref} \) and the LR image \( \text{LR} \), Ref is processed via bicubic downsampling-upsampling to obtain \( \text{Ref}\downarrow\uparrow \), while LR is upsampled to generate \(\text{LR}\uparrow \). The preprocessed images and the original HR Ref are fed into the MFAM in parallel to obtain three multi-scale deep features $f_\text{MFAM}^s(\text{LR}\uparrow)$, $f_\text{MFAM}^s(\text{LR}\uparrow)$ and $f_\text{MFAM}^s(\text{LR}\uparrow)$, here $s\in\{1,2,3\}$. The preprocessed dual-modal data and the original HR Ref are fed into the MFAM. The output features are unfolded into multiple patches and vectorized, resulting in vector shapes of ${Q}_i^s\in[1,H_\text{LR↑}^s\times W_\text{LR↑}^s]$, ${K}_i^s\in[1,H_\text{Ref↓↑}^s\times W_\text{Ref↓↑}^s]$ and ${V}_i^s\in[1,H_\text{Ref}^s\times W_\text{Ref}^s]$, which are concatenated to form the final ${Q^s}$, ${K^s}$, and ${V^s}$.
\begin{align*}\tag{16}
\left\{
\begin{aligned}
    Q^s &= f_\text{unfold}(f_\text{MFAM}^s(\text{LR}\uparrow))\\
    K^s &= f_\text{unfold}(f_\text{MFAM}^s(\text{Ref}\downarrow\uparrow))\\
    V^s &= f_\text{unfold}(f_\text{MFAM}^s(\text{Ref}))\\
\end{aligned}
\right.
\end{align*}

\subsubsection{Local-Global Correlation Modeling}
Local (${R}_{\text{local}}^s$) and global (${R}_{\text{global}}^s$) correlation matrices are constructed using cosine similarity, measuring neighborhood texture similarity and cross-region semantic relationships, respectively. These are adaptively fused via a learnable coefficient $\alpha$.
\begin{align*}\tag{17}
\begin{aligned}
    &{R}_{\text{local}}^s = \text{Norm}({Q}^s) \cdot \text{Norm}({K}^s)^T\\
    &{R}_{\text{global}}^s = \text{Norm}({\bar{Q^s}}) \cdot \text{Norm}({\bar{K^s}})^T\\
    &{R}_{\text{mix}}^s = \alpha {R}_{\text{global}}^s + (1-\alpha){R}_{\text{local}}^s  
\end{aligned}
\end{align*}

\subsubsection{Dynamic Top-K Sparsification}
To adaptively determine the optimal sparsity level for each hierarchy, a Gumbel-Softmax-based probability distribution and Straight-Through Estimator (STE\cite{STE}) are introduced. Discrete Top-K selection is used in forward propagation, while continuous probability gradients are approximated during backpropagation.
\begin{align*}\tag{18}
\begin{aligned}
{P} &= \text{Gumbel-Softmax}({W}_k) \\  
K &= \arg\max({P}) + \left(\sum_{k=1}^{K_{\max}} {P}_k \cdot k \right) - \arg\max({P})  
\end{aligned}  
\end{align*}
Among them, ${W}_k$ is the learnable parameter, $K_{\text{max}}$ is the maximum sparsity.

\subsubsection{Soft-Hard Attention Collaborative Transfer}
Target-oriented texture features are extracted through pointwise multiplication between the sparse correlation matrix $ {R}_\text{sparse}^s$ and the index map ${I}^s$. The corresponding texture details are obtained by multiplying the sparse index ${I}^s$ with ${V}^s$. These details are then weighted by multiplying with the sparsity-related matrix $ {R}_\text{sparse}^s$ to obtain the final sparse texture guidance information ${T}^s$.
\begin{align*}\tag{19}
    \begin{aligned}  
    {V}'_s = {V}^s \odot {I}^s, \quad {T}^s = {R}_{\text{sparse}}^s \odot {V}'_s  
    \end{aligned}  
\end{align*}
\subsection{Loss Function}
In our algorithm, there are three loss functions in total. The overall loss function can be interpreted as:
\begin{equation}\tag{20}
    L = \lambda_1 \times L_{diff} + \lambda_2 \times L_{\text{pixel}} + \lambda_3 \times L_{\text{per}}
\end{equation}

We set the weights of the three loss functions as: $\lambda_1=1$, $\lambda_2=10^{-3}$, $\lambda_3=10^{-4}$. Below, we specifically introduce the specific meanings of each loss function.

\subsubsection{Diffusion Loss}
The objective function of the diffusion model is to minimize the KL divergence between the forward diffusion process $q(I_{t-1}\vert I_t,I_0,I_{\text{res}})$ and the backward inference $p_\theta(I_{t-1}\vert I_t)$. Similar to DDPM, we use the following formula to calculate the loss function between the predicted residual and the true residual: 
\begin{equation}\tag{21}
    L_{\text{res}}(\theta) = \mathbb{E}[\lambda_{res}\| I_{\text{res}} - I_{\text{res}}^\theta(I_t,t,I_{\text{in}}) \|^2]
\end{equation}

the loss function between the predicted noise and the true noise:
\begin{equation}\tag{22}
    L_{\varepsilon}(\theta) = \mathbb{E}[\lambda_{\varepsilon}\| \varepsilon - \varepsilon_{\theta}(I_t,t,I_{\text{in}}) \|^2]
\end{equation}

Here, $\lambda_{res}, \lambda_{\varepsilon} \in [0,1]$, used for re-weighting. We set \(\lambda_{res} = 1\), \(\lambda_{\varepsilon} = 1\), and the final diffusion loss is:
\begin{equation}\tag{23}
    L_{\text{diff}}(\theta) =  L_{\text{res}}(\theta) + L_{\varepsilon}(\theta) 
\end{equation}

\subsubsection{Pixel Loss}
We derive the predicted noise$I_\varepsilon^\theta$ and residual$I_{\text{res}}^\theta$ through the reverse inference process, and estimate the original clean image at each time step based on Equation \ref{eq12}. We use the L1 pixel loss to calculate the pixel error between SR($\hat{I}_\theta$) and HR($I_0$), with the specific formula being:
\begin{equation}\tag{24}
    L_{\text{pixel}}(\theta) = \mathbb{E}[ \ \| \hat{I}_{ijk}^\theta - I_{ijk}^0 \|_1]
\end{equation}
By introducing pixel loss, we ensure that the SR image conforms to the Ground Truth.

\subsubsection{Perceived Loss}
The core idea of perceptual loss is to compute the difference between two images using a pre-trained neural network, thereby enhancing the similarity between the SR image and the HR image in the deep feature space. We use a pre-trained VGG network to extract deep features, and the perceptual loss can be interpreted as:
\begin{equation}\tag{25}
    L_{\text{per}}(\theta) = \mathbb{E}[\ \| \phi_{\text{VGG}}(\hat{I}_{ijk}^\theta) - \phi_{\text{VGG}}(I_{ijk}^0) \|^2]
\end{equation}

Here, $\phi_{\text{VGG}}$ denotes the feature maps of the fourth layer of VGG19. This loss function can minimize the distance between SR and HR in the feature space, thereby enriching the details of the SR image.

\begin{algorithm*}[t!]
\caption{Training of TTRD3}
\label{algtrain}
\SetKwInOut{Input}{Input}
\SetKwInOut{Initialization}{Initialization}
\Input{HR image $I_0$, upsampled LR image $I_{\text{in}}$, Ref image $I_{\text{ref}}$, Downsampled-Upsampled Ref image $I_{\text{refsr}}$, total step $T$}
\Initialization{$\epsilon \sim \mathcal{N}(0, \delta^2)$, $t\in[0,T]$, $I_{\text{res}} = I_{\text{in}} - I_0$, diffusion network $\psi_{\theta}$, $\bar{\alpha}_t = \sum_{i=1}^t \alpha_i$, $\quad \bar{\beta}_t = \sqrt{\sum_{i=1}^t \beta_i^2}$}
\Repeat{
    converged}{
    $Q = f_{\text{MFAM}}(I_{\text{in}})$, $K = f_{\text{MFAM}}(I_{\text{refsr}})$, $V = f_{\text{MFAM}}(I_{\text{ref}})$ \hfill{// Multi-scale Feature Aggregation}
    
    $T_{\text{guid}} = f_{\text{STTG}}(Q, K, V)$ \hfill{// Sparse Texture Transfer Guidance}
    
    $I_t = I_0 + \bar{\alpha}_t I_{\text{res}} + \bar{\beta}_t \epsilon$ 
    
    $\epsilon_{\theta}, I_{\text{res}}^{\theta} = \psi_{\theta}(I_t, I_{\text{in}}, T_{\text{guid}})$ \hfill{// Predict noise and residual}
    
    $I_0 = I_t - \bar{\alpha}_t I_{\text{res}}^{\theta} - \bar{\beta}_t \epsilon_{\theta}$ \hfill{// Reconstruct HR image}
    
    $\mathcal{L}_{\text{overall}} = \lambda_1 \mathcal{L}_{\text{diffusion}} + \lambda_2 \mathcal{L}_{\text{pixel}} + \lambda_3 \mathcal{L}_{\text{perception}}$ \hfill{// Loss calculation}
    
    Update $\theta$ via $\nabla_{\theta} \mathcal{L}_{\text{overall}}$ \hfill{// Gradient descent}
    }
\end{algorithm*}

\begin{algorithm*}[t!]
\caption{Inference of TTRD3}
\label{alginfer}
\SetKwInOut{Input}{Input}
\SetKwInOut{Output}{Output}
\SetKwInOut{Initialization}{Initialization}
\SetKwInOut{Inference}{Inference}
\Input{Upsampled LR image $I_{\text{in}}$, Ref image $I_{\text{ref}}$, total step $T$, Sampling steps $T_{\text{sample}}$, Downsampled-Upsampled Ref image $I_{\text{refsr}}$}
\Output{SR image $I_{\text{sr}}$}
\Initialization{$\varepsilon \sim \mathcal{N}(0, \delta^2)$, Degrade image: $I_T = I_{\text{in}} + \bar{\beta}_T \varepsilon$, diffusion network $\psi_{\theta}$}
\Inference{}

    $Q = f_{\text{MFAM}}(I_{\text{in}})$ , $K = f_{\text{MFAM}}(I_{\text{refsr}})$, $V = f_{\text{MFAM}}(I_{\text{ref}})$ \hfill{// Multi-scale Feature Aggregation}
    
    $T_{\text{guid}} = f_{\text{STTC}}(Q, K, V)$ \hfill{// Sparse Texture Transfer Guidance}

    \For{$t =T$ \textbf{down to} $1$ \textbf{step} $\left\lfloor T/(T_{\text{sample}} - 1) \right\rfloor$}
    {
        $\alpha_t = \sum_{i=1}^t \alpha_i$, $\quad \bar{\beta}_t = \sqrt{\sum_{i=1}^t \beta_i^2}$
        
        $\varepsilon_{\theta}, I_{\text{res}}^{\theta} = \psi_{\theta}(I_t, I_{\text{in}}, T_{\text{guid}})$ \hfill{// Predict noise and residual}
        
        $I_{\text{prev}} = I_t - (\alpha_t - \alpha_{\text{prev}})I_{\text{res}}^{\theta} - (\bar{\beta}_t - \sqrt{\bar{\beta}_{\text{prev}}^2 - \sigma_t^2})\varepsilon_{\theta} + \sigma_t \varepsilon$ 
    }

    $I_{\text{sr}} = \hat{I}_0$ \hfill{// Predic SR image}

\end{algorithm*}

\section{Experimental Details}\label{sec4}
\subsection{Dataset}

We use public RS datasets AID and RSD46 to comprehensively evaluate the effectiveness of the proposed method. The AID dataset contains 10,000 images across 30 scene categories, each with a size of $600\times600$ pixels. The RSD46 dataset includes 117,000 images from 46 scene categories, each sized $256\times256$ pixels.

To enhance network generalization and reduce computational costs, we crop each image in the AID dataset into four $256\times256$ patches and randomly select one patch for use. For the RSD46 dataset, 100 images are randomly selected from each of the 46 scene categories. Specifically, a total of 14,600 images are divided into training, reference, and validation sets in a $[6:3:1]$ ratio. Consequently, the training set contains 8,760 images, the reference image set contains 4,380 images, and the validation set includes 1,460 images. All data are degraded using bicubic interpolation to construct $4\times$ SR tasks.

\subsection{Evaluation Metrics}
We use four metrics for SR image quality assessment. Among these, PSNR is calculated as the ratio between the maximum signal power and noise power based on Mean Square Error. SSIM\cite{SSIM} measures local structural similarity between images through three dimensions: luminance, contrast, and structure. LPIPS\cite{LPIPS} is a perceptual similarity metric based on deep feature space, designed to compute weighted distances in the feature space. FID\cite{FID} evaluates the realism and diversity of generated images by comparing the distribution differences between generated and real images in the Inception-v3 feature space.  

In summary, PSNR and SSIM focus on pixel-level and structural reconstruction fidelity, whereas LPIPS and FID emphasize human visual perception-based quality assessment. The combined use of these four metrics provides a comprehensive evaluation of algorithmic reconstruction accuracy and perceptual quality.

\begin{table*}[t!]
  \centering
  \caption{The quantitative LPIPS comparison with SOTA SR models on 30 scene categories of the AID test set. the best LPIPS value in each category is highlighted in {\textcolor{red}{RED}}, while the second best is in {\textcolor{blue}{BLUE}}.}
    \resizebox{1\textwidth}{!}{
    \begin{tabular}{cccccccccccc}
    \noalign{\hrule height 1.5pt} 
    \multicolumn{1}{c}{\multirow{2}{*}{Categories}} & 
    \multicolumn{1}{c}{} & \multicolumn{2}{c}{CNN-based} & 
    \multicolumn{2}{c}{Transformer-based} & \multicolumn{2}{c}{GAN-based} & 
    \multicolumn{4}{c}{Diffusion-based} \\
    & {Bicubic} & {EDSR} & {RCAN} & {SwinIR} & {TTST} & {ESRGAN} & {LDL} & {SR3} & {EDiffSR} & {ResShift} & {TTRD3} \\
    \hline
    airport          & 0.3620 & 0.3032 & 0.3041 & 0.3036 & 0.2994 & 0.2814 & {\textcolor{blue}{0.2683}} & 0.3572 & 0.3844 & 0.2736 & {\textcolor{red}{0.2655}} \\
    bareland         & 0.2817 & 0.2568 & 0.2554 & 0.2568 & 0.2540 & 0.2412 & {\textcolor{red}{0.2225}} & 0.2660 & 0.4299 & 0.3234 & {\textcolor{blue}{0.2271}} \\
    baseballfield    & 0.2694 & 0.2375 & 0.2356 & 0.2373 & 0.2340 & 0.2322 & {\textcolor{blue}{0.2203}} & 0.2893 & 0.4080 & 0.2465 & {\textcolor{red}{0.2142}} \\
    beach            & 0.3529 & 0.3285 & 0.3274 & 0.3270 & 0.3235 & 0.2893 & {\textcolor{blue}{0.2776}} & 0.3402 & 0.3898 & 0.2876 & {\textcolor{red}{0.2709}} \\
    bridge           & 0.2923 & 0.2449 & 0.2449 & 0.2443 & 0.2429 & 0.2304 & {\textcolor{blue}{0.2185}} & 0.3174 & 0.3671 & 0.2392 & {\textcolor{red}{0.2114}} \\
    center           & 0.3551 & 0.2781 & 0.2772 & 0.2782 & 0.2724 & 0.2659 & 0.2652 & 0.3764 & 0.3573 & {\textcolor{blue}{0.2494}} & {\textcolor{red}{0.2447}} \\
    church           & 0.3942 & 0.3342 & 0.3335 & 0.3348 & 0.3314 & 0.3070 & {\textcolor{blue}{0.2933}} & 0.3912 & 0.3626 & 0.3024 & {\textcolor{red}{0.2848}} \\
    commercial       & 0.3755 & 0.3192 & 0.3199 & 0.3216 & 0.3173 & 0.2982 & {\textcolor{blue}{0.2903}} & 0.3886 & 0.3750 & 0.2931 & {\textcolor{red}{0.2836}} \\
    denseresidential & 0.4299 & 0.3569 & 0.3590 & 0.3570 & 0.3533 & 0.3066 & {\textcolor{blue}{0.3014}} & 0.4106 & 0.3570 & 0.3149 & {\textcolor{red}{0.2885}} \\
    desert           & 0.2865 & 0.2523 & 0.2522 & 0.2520 & 0.2488 & 0.2281 & {\textcolor{red}{0.2157}} & 0.3003 & 0.4145 & 0.2617 & {\textcolor{blue}{0.2228}} \\
    farmland         & 0.3430 & 0.2948 & 0.2921 & 0.2959 & 0.2936 & 0.2745 & {\textcolor{blue}{0.2655}} & 0.3645 & 0.4246 & 0.2820 & {\textcolor{red}{0.2614}} \\
    forest           & 0.4092 & 0.4031 & 0.4028 & 0.4040 & 0.4027 & {\textcolor{blue}{0.3196}} & 0.3248 & 0.3482 & 0.3803 & 0.3279 & {\textcolor{red}{0.3059}} \\
    industrial       & 0.3369 & 0.2669 & 0.2669 & 0.2654 & 0.2606 & 0.2529 & 0.2377 & 0.3486 & 0.3165 & {\textcolor{blue}{0.2360}} & {\textcolor{red}{0.2324}} \\
    meadow           & 0.3489 & 0.3392 & 0.3362 & 0.3376 & 0.3338 & 0.3094 & {\textcolor{blue}{0.2968}} & 0.3283 & 0.4319 & 0.4137 & {\textcolor{red}{0.2863}} \\
    mediumresidential& 0.3749 & 0.3338 & 0.3339 & 0.3334 & 0.3284 & 0.2771 & 0.2745 & 0.3379 & 0.3261 & {\textcolor{blue}{0.2726}} & {\textcolor{red}{0.2562}} \\
    mountain         & 0.3680 & 0.3486 & 0.3514 & 0.3491 & 0.3487 & 0.3238 & {\textcolor{blue}{0.3102}} & 0.3614 & 0.4183 & 0.3132 & {\textcolor{red}{0.3071}} \\
    park             & 0.3496 & 0.3093 & 0.3094 & 0.3086 & 0.3074 & 0.2896 & 0.2849 & 0.3537 & 0.3829 & {\textcolor{blue}{0.2792}} & {\textcolor{red}{0.2725}} \\
    parking          & 0.3094 & 0.2316 & 0.2298 & 0.2297 & {\textcolor{blue}{0.2264}} & 0.2411 & 0.2271 & 0.3472 & 0.3018 & 0.2305 & {\textcolor{red}{0.2089}} \\
    playground       & 0.3098 & 0.2510 & 0.2490 & 0.2502 & 0.2482 & 0.2308 & {\textcolor{blue}{0.2177}} & 0.3090 & 0.3562 & 0.2497 & {\textcolor{red}{0.2134}} \\
    pond             & 0.2750 & 0.2536 & 0.2537 & 0.2556 & 0.2583 & 0.2332 & {\textcolor{blue}{0.2302}} & 0.3385 & 0.3940 & 0.2471 & {\textcolor{red}{0.2162}} \\
    port             & 0.3026 & 0.2600 & 0.2611 & 0.2604 & 0.2584 & 0.2447 & {\textcolor{blue}{0.2335}} & 0.3475 & 0.3690 & 0.2395 & {\textcolor{red}{0.2253}} \\
    railwaystation   & 0.3812 & 0.3069 & 0.3074 & 0.3075 & 0.3012 & 0.2867 & {\textcolor{blue}{0.2702}} & 0.3849 & 0.3521 & 0.2743 & {\textcolor{red}{0.2653}} \\
    resort           & 0.3209 & 0.2744 & 0.2753 & 0.2752 & 0.2711 & 0.2574 & {\textcolor{blue}{0.2449}} & 0.3238 & 0.3345 & 0.2472 & {\textcolor{red}{0.2427}} \\
    river            & 0.3481 & 0.3202 & 0.3204 & 0.3201 & 0.3169 & 0.2785 & 0.2747 & 0.3357 & 0.3716 & {\textcolor{blue}{0.2734}} & {\textcolor{red}{0.2598}} \\
    school           & 0.3915 & 0.3317 & 0.3330 & 0.3338 & 0.3286 & 0.2981 & {\textcolor{blue}{0.2907}} & 0.3846 & 0.3642 & 0.3009 & {\textcolor{red}{0.2815}} \\
    sparseresidential& 0.4050 & 0.3938 & 0.3919 & 0.3924 & 0.3880 & 0.3073 & 0.3125 & 0.3327 & 0.3367 & {\textcolor{blue}{0.3059}} & {\textcolor{red}{0.2911}} \\
    square           & 0.2921 & 0.2383 & 0.2371 & 0.2386 & 0.2337 & 0.2182 & {\textcolor{blue}{0.2070}} & 0.2975 & 0.3183 & 0.2117 & {\textcolor{red}{0.2042}} \\
    stadium          & 0.3133 & 0.2502 & 0.2492 & 0.2513 & 0.2453 & 0.2449 & {\textcolor{blue}{0.2307}} & 0.3224 & 0.3331 & 0.2376 & {\textcolor{red}{0.2225}} \\
    storagetanks     & 0.3765 & 0.3239 & 0.3247 & 0.3251 & 0.3229 & 0.2862 & {\textcolor{blue}{0.2750}} & 0.3555 & 0.3493 & 0.2758 & {\textcolor{red}{0.2695}} \\
    viaduct          & 0.3703 & 0.3065 & 0.3044 & 0.3061 & 0.3057 & 0.2772 & 0.2703 & 0.3465 & 0.3411 & {\textcolor{blue}{0.2611}} & {\textcolor{red}{0.2564}} \\
    \hline
    AVERAGE & 0.3442 & 0.2983 & 0.2980 & 0.2984 & 0.2952 & 0.2711 & {\textcolor{blue}{0.2624}} & 0.3435 & 0.3683 & 0.2757 & {\textcolor{red}{0.2531}} \\
    \noalign{\hrule height 1.5pt} 
    \end{tabular}%
    }
  \label{tab1}%
\end{table*}%

\subsection{Implementation Details}
Our study focuses on $4\times$ SR. In the final TTRD3 architecture, the number of MFABs in the MFAM is set to $[2, 2, 2]$ to achieve cross-scale feature fusion. For the residual denoising dual diffusion network, we employ a U-Net with an input channel number of $C=64$, four depth levels, and channel multipliers of $[1, 2, 4, 8]$ at each depth. The TTRD3-2Net network architecture was employed. The noise intensity added is $\bar{\beta}_T=0.1$. The preset maximum K value is 10.

To train the TTRD3 network, we perform 300k iterations with a batch size of 2. The learning rate is initialized to $1\times{10}^{-4}$, with diffusion loss weight set to $1$, pixel loss weight to $1\times{10}^{-3}$, and perceptual loss weight to $1\times{10}^{-4}$. Optimization is conducted using the Adam optimizer with $\beta_1=0.9$ and $\beta_2=0.99$. The diffusion process uses a total timestep $T=1000$ and $10$ sampling steps.

All comparative methods are retrained on our dataset using their official code implementations and hyperparameters to ensure fair comparison. Experiments are conducted on a single NVIDIA RTX 4090 GPU (24GB memory) with the PyTorch framework.

The pseudocode for the training and inference processes of the proposed TTRD3 algorithm can be found in Algorithm \ref{algtrain} and Algorithm \ref{alginfer}, respectively.

\begin{table*}[t!]
  \centering
  \caption{The quantitative comparison with SOTA SR models on the AID and RSD46 test sets is presented, with the best performance of the same model networks highlighted in \textbf{bold}.}
   \resizebox{1\textwidth}{!}{
    \begin{tabular}{ccc|cc|cc|cc|cccc}
    \noalign{\hrule height 1.5pt} 
    \multirow{2}[2]{*}{Dataset} & \multirow{2}[2]{*}{Metrics} & Baseline & \multicolumn{2}{c|}{CNN-based} & \multicolumn{2}{c|}{Transformer-based} & \multicolumn{2}{c|}{GAN-based} & \multicolumn{4}{c}{Diffusion-based} \bigstrut[t]\\
          &       & Bicubic & EDSR  & RCAN  & SwinIR & TTST  & ESRGAN & LDL   & SR3   & EDiffSR & ResShift & TTRD3 \bigstrut[b]\\
    \hline
    \multirow{4}[2]{*}{AID} & PSNR(↑) & 27.9757  & 29.9272  & \textbf{30.1317 } & 30.1144  & \textbf{30.1627 } & 27.4881  & \textbf{27.7927 } & 26.4684  & 25.2836  & 28.1733  & \textbf{28.6798} \bigstrut[t]\\
          & SSIM(↑) & 0.7298  & 0.8117  & \textbf{0.8156 } & 0.8156  & \textbf{0.8172 } & 0.7377  & \textbf{0.7533 } & 0.6921  & 0.5946  & 0.7498  & \textbf{0.7679 } \\
          & LPIPS(↓) & 0.3442  & 0.2983  & \textbf{0.2980 } & 0.2984  & \textbf{0.2952 } & 0.2711 & \textbf{0.2624 }  & 0.3435  & 0.3683  & 0.2757  & \textbf{0.2531 } \\
          & FID(↓) & 85.7452  & 79.9425  & \textbf{79.8817 } & 80.6497  & \textbf{78.0137 } & 35.4819  & \textbf{34.9917 } & 72.2483  & 54.0955  & 54.8146  & \textbf{33.6997 } \bigstrut[b]\\
    \hline
    \multirow{4}[2]{*}{RSD46} & PSNR(↑) & 27.9253  & 30.0547  & \textbf{30.2306 } & 30.2271  & \textbf{30.2320 } & 27.6221  & \textbf{27.8908 } & 26.5481  & 25.2777  & 28.5018  & \textbf{28.8727 } \bigstrut[t]\\
          & SSIM(↑) & 0.7364  & 0.8267  & \textbf{0.8302 } & 0.8307  & \textbf{0.8308 } & 0.7480  & \textbf{0.7522 } & 0.6982  & 0.6024  & 0.7642  & \textbf{0.7782 } \\
          & LPIPS(↓) & 0.3310  & 0.2733  & \textbf{0.2724 } & 0.2732  & \textbf{0.2721 } & \textbf{0.2450 } & 0.2478  & 0.3277  & 0.3596  & 0.2543  & \textbf{0.2442 } \\
          & FID(↓) & 102.7406  & 100.6739  & \textbf{99.6016 } & 100.9893  & \textbf{99.1222 } & \textbf{50.5132 } & 51.5891  & 93.3110  & 78.8586  & 83.7268  & \textbf{48.7845 } \bigstrut[b]\\
      \noalign{\hrule height 1.5pt} 
    \end{tabular}%
    }
  \label{tab2}%
\end{table*}%

\section{Experiments and Discussion}\label{sec5}
\subsection{Comparison with SOTA Results}
We compare our TTRD3 with SOTA SR methods, including EDSR\cite{EDSR}, RCAN\cite{RCAN}, SwinIR\cite{Swinir}, ESRGAN\cite{ESRGAN} , LDL\cite{LDL}, SR3\cite{SR3}, ResShift\cite{Resshift}, and two RS-specific SOTA methods: EDiffSR\cite{Ediffsr} and TTST\cite{TTST}. These methods represent mainstream approaches in the field: EDSR and RCAN are CNN-based SR networks; SwinIR and TTST are Transformer-based; ESRGAN and LDL are GAN-based; SR3, EDiffSR, and ResShift are DM-based. All methods are retrained on our dataset using their official configurations for fair evaluation.
\subsubsection{Quantitative Comparison}

Table \ref{tab1} shows the LPIPS results across 30 scene categories in the AID test set. The best and second-best LPIPS scores are highlighted in each row. Our TTRD3 achieves superior LPIPS performance in most scenarios,  demonstrating robust generalization across diverse RS scenes.

Table \ref{tab2} summarizes the average PSNR, SSIM, LPIPS, and FID scores on the AID and RSD46 test sets. TTRD3 achieves the best LPIPS and FID scores while maintaining competitive PSNR and SSIM metrics. It is critical to emphasize that while PSNR is highly correlated with pixel-level differences and SSIM can reflect the structural similarity of images, images with high PSNR and SSIM values do not always yield perceptually satisfactory results. CNN-based and Transformer-based networks exhibit superior PSNR and SSIM scores, but their overly smoothed reconstructions fail to align with human perceptual expectations\cite{r7}. In contrast, GAN-based and DM-based SR methods demonstrate competitive performance in perceptual metrics (LPIPS/FID), delivering more visually realistic outputs. Notably, TTRD3 not only attains the highest LPIPS and FID scores but also maintains optimal PSNR and SSIM performance among GAN-based and DM-based approaches, unequivocally validating its exceptional SR reconstruction capabilities.

\begin{figure*}[t!]
    \centering
    \includegraphics[width=1\textwidth]{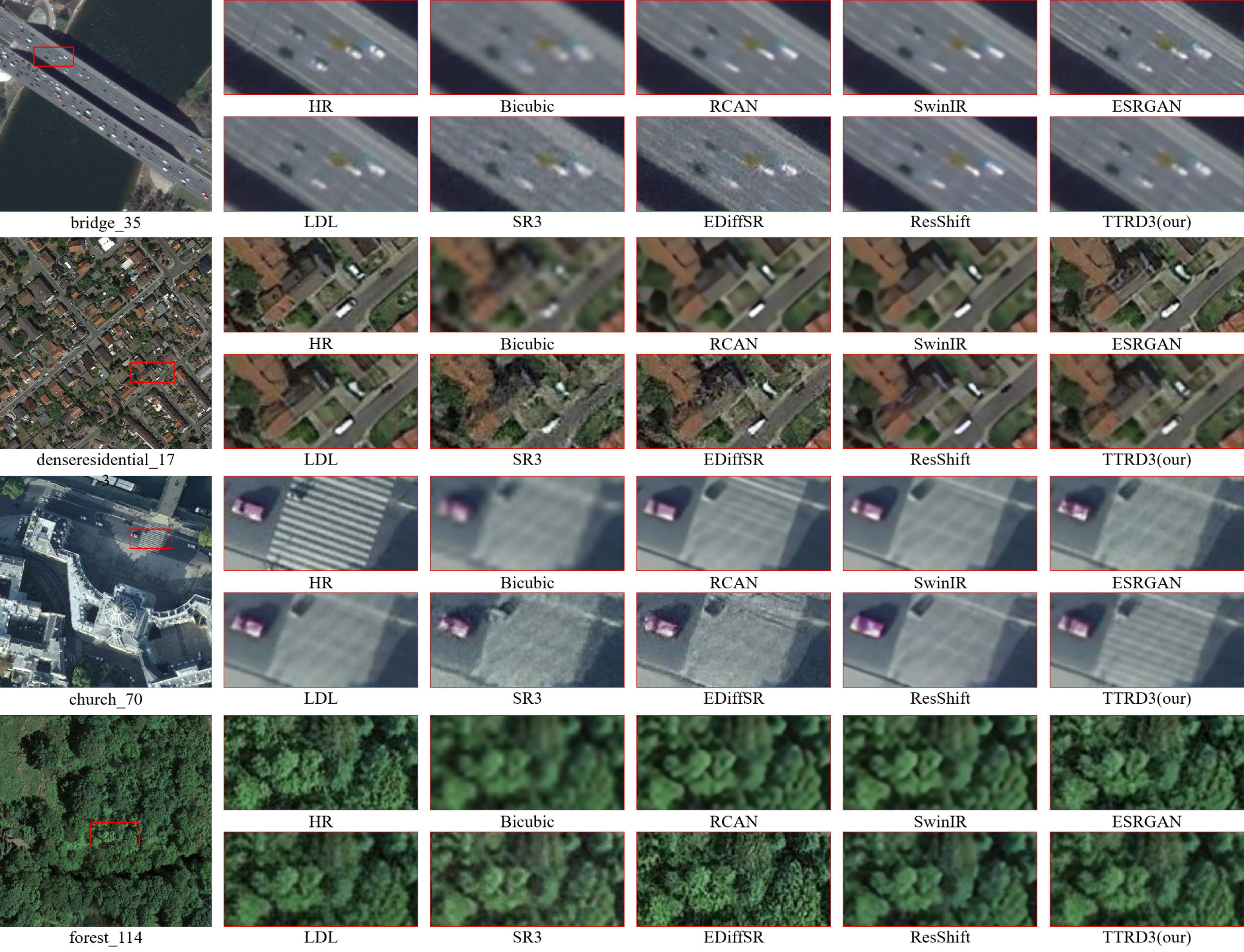}
    \caption{The visual comparison of the $4\times$ SR task on the AID test set with other SOTA SR models. Zoom in for a better view.}
    \label{figAID}
\vspace{10pt}
\end{figure*}

\begin{figure*}[t!]
    \includegraphics[width=1\textwidth]{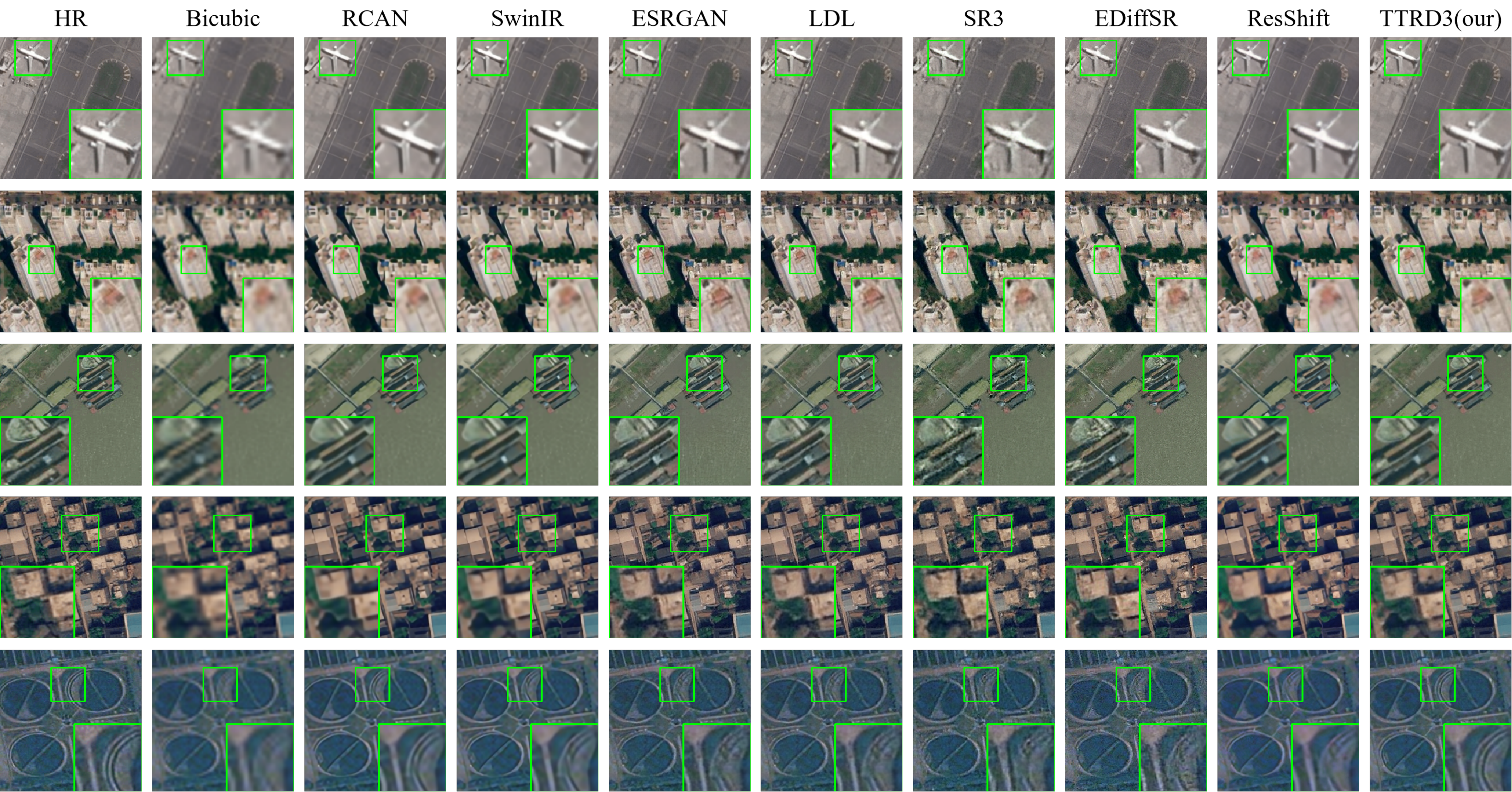}
    \caption{The visual comparison of the $4\times$ SR task on the RSD46 test set with other SOTA SR models. Zoom in for a better view.}
    \label{figRSD46}
\end{figure*}

\subsubsection{Qualitative Comparison}
Our TTRD3 is visually compared with other models. As shown in Figure \ref{figAID} from the AID dataset, the proposed TTRD3 demonstrates the best visual authenticity and accuracy. For the ``forest\textunderscore114'' image, regression-based methods (RCAN, SwinIR) produce over-smoothed results, while GAN-based methods (ESRGAN, LDL) exhibit severe artifacts with repetitive structures inconsistent with ground truth, despite their realistic appearance. DM-based methods (SR3, EDiffSR, ResShift) struggle to balance authenticity and accuracy. In contrast, our TTRD3, which reconstructs images through dual residual-noise diffusion, achieves highly realistic details while maintaining superior accuracy, demonstrating its powerful capability to generate visually appealing results.

For the ``church\textunderscore70'' image, the over-smoothed outputs of RCAN and SwinIR fail to restore regular high-frequency texture like zebra crossings. Generative models such as ESRGAN, LDL, SR3, EDiffSR, and ResShift also struggle to recover sufficient details due to the lack of high-frequency guidance. TTRD3 addresses this by adaptively introducing structurally similar reference textures from HR images, effectively enhancing spatial high-frequency restoration and improving the performance of the dual diffusion model in SR tasks.

 As shown in Figure \ref{figRSD46} from the RSD46 dataset, regression-based methods (RCAN, SwinIR) retain reasonable accuracy in reconstructing regular structures but yield blurred visuals. GAN-based and diffusion-based methods (ESRGAN, LDL, SR3, EDiffSR, ResShift) generate richer high-frequency details but introduce significant structural distortions. TTRD3 not only produces photorealistic high-frequency textures but also accurately reconstructs regular structures. As demonstrated in Figure \ref{figRSD46}, its reconstruction quality for targets such as aircraft, ships, and buildings significantly outperforms competing methods, proving that TTRD3 achieves an exceptional balance between authenticity and accuracy to deliver realistic and precise results.

\begin{table}[t!]
  \centering
  \caption{The ablation study results of each module in TTRD3 on the AID test set.}
   \resizebox{0.48\textwidth}{!}{
    \begin{tabular}{>{\centering\arraybackslash}p{3.5em}
    >{\centering\arraybackslash}p{2.3em}
    >{\centering\arraybackslash}p{2.3em}
    >{\centering\arraybackslash}p{2.3em}
    >{\centering\arraybackslash}p{2.3em}
    >{\centering\arraybackslash}p{2.8em}
    >{\centering\arraybackslash}p{2.8em}
    >{\centering\arraybackslash}p{2.8em}}
      \noalign{\hrule height 1.5pt} 
      Methods & DDPM  & RDDM  & STTG  & MFAM  & PSNR↑ & LPIPS↓ & FID↓ \bigstrut\\
      \midrule 
      Baseline & \checkmark     & ×     & ×     & ×     & 26.4684 & 0.3435 & 72.2483 \bigstrut[t]\\
      Model-1 & ×     & \checkmark     & ×     & ×     & 28.2994 & 0.2675 & 36.4484 \\
      Model-2 & ×     & \checkmark     & \checkmark     & ×     & 28.5099 & 0.2607 & 34.0630 \\
      TTRD3 & ×     & \checkmark     & \checkmark     & \checkmark     & 28.6798 & 0.2531 & 33.6997 \bigstrut[b]\\
      \noalign{\hrule height 1.5pt}
    \end{tabular}%
    }
  \label{tab3}%
\end{table}

\begin{figure}[t!]  
    \centering  
    \includegraphics[width=0.49\textwidth]{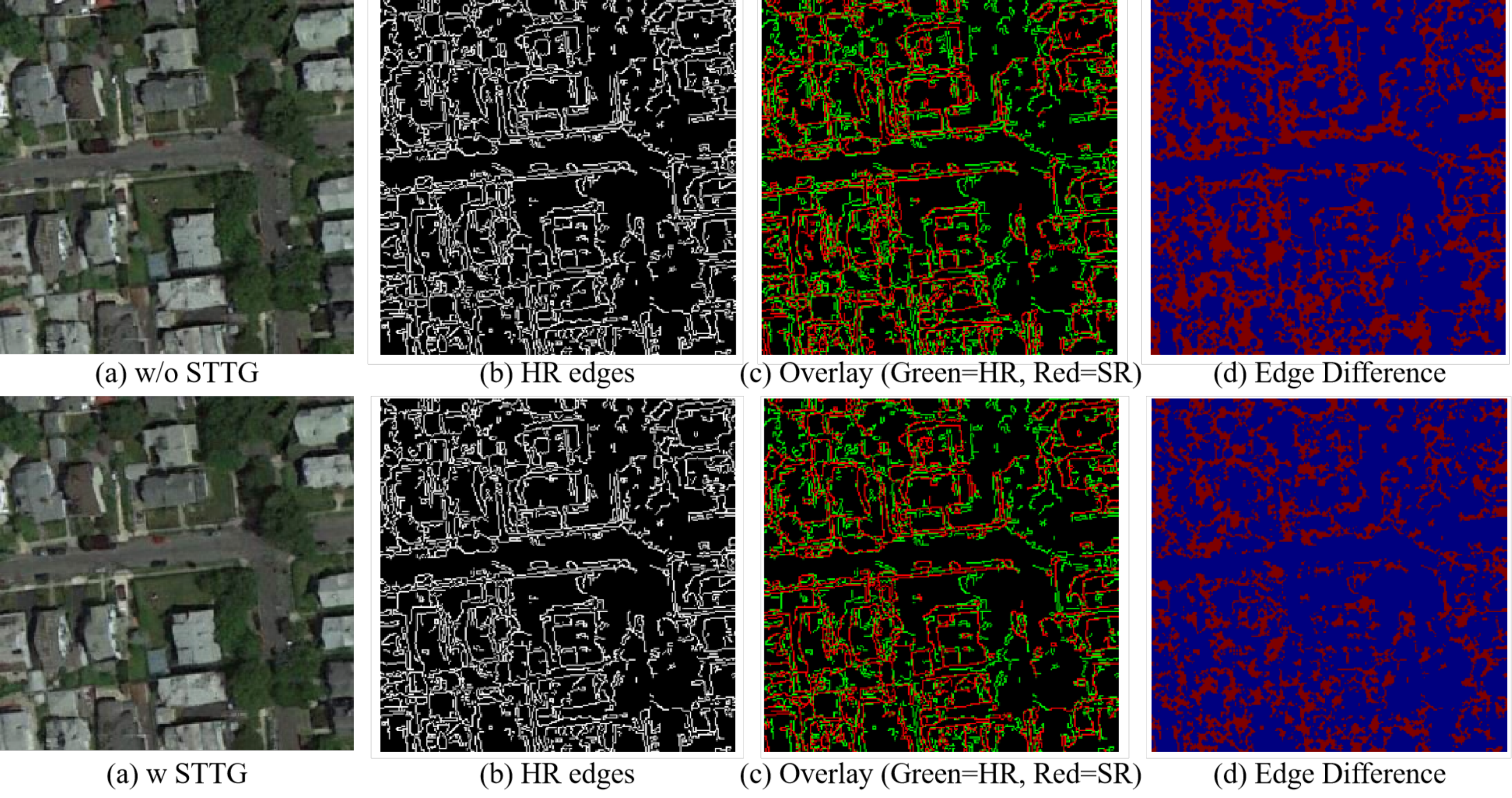}
    \caption{Visualization of ablation comparisons for the STTG module: (a) SR results with and without the STTG module, (b) boundary maps of HR images obtained using the Canny operator, (c) overlay comparison between HR and SR boundaries, and (d) visualized boundary discrepancies.}  
    \label{figwosttg}  
\end{figure}

\subsection{Ablation Study}
In this section, we conduct extensive experiments to validate the effectiveness of each component in our TTRD3 framework. As shown in Table \ref{tab3}.

\subsubsection{Effectiveness of the RDDM}
To validate the effectiveness of the RDDM in SR tasks, we remove the MFAM and STTG, using only the LR image and the degraded image $I_T$ (via channel concatenation) as network inputs. Experimental results show that Model-1 significantly outperforms the baseline model relying solely on noise diffusion across all metrics (PSNR: +1.8310 dB, LPIPS: -0.0760, FID: -35.7999). This confirms that residual information enhances reconstruction fidelity, while noise diffusion optimizes generation quality.

\subsubsection{Effectiveness of the STTG}
To evaluate the contribution of STTG, we integrate it into the RDDM framework and replace the MFAB in MFAM with standard residual blocks, constructing Model-2. Comparative experiments demonstrate that Model-2 achieves superior performance over Model-1 across all metrics (PSNR: +0.2105 dB, LPIPS: -0.0068, FID: -2.3854). Figure \ref{figwosttg} demonstrates the ablation study results of the STTG.This proves that STTG effectively extracts sparse texture guidance from reference images and significantly enhances the SR generation performance of the dual diffusion model.

\subsubsection{Effectiveness of the MFAM} 
By comparing the full TTRD3 framework with Model-2, we observe that TTRD3 achieves significant advantages in PSNR, SSIM, LPIPS, and FID (PSNR: +0.1699 dB, LPIPS: -0.0076, FID: -0.3633). This validates the critical role of MFAB in extracting multi-scale feature information from RS images, providing essential supplementary information for STTG's sparse texture extraction, thereby further improving SR performance.

\subsection{Analysis of the Top-K selection Mechanism in STTG}
\subsubsection{Trade-off Analysis between Noise Suppression and Detail Preservation}

\begin{figure}[t!]  
    \centering
    \includegraphics[width=0.49\textwidth]{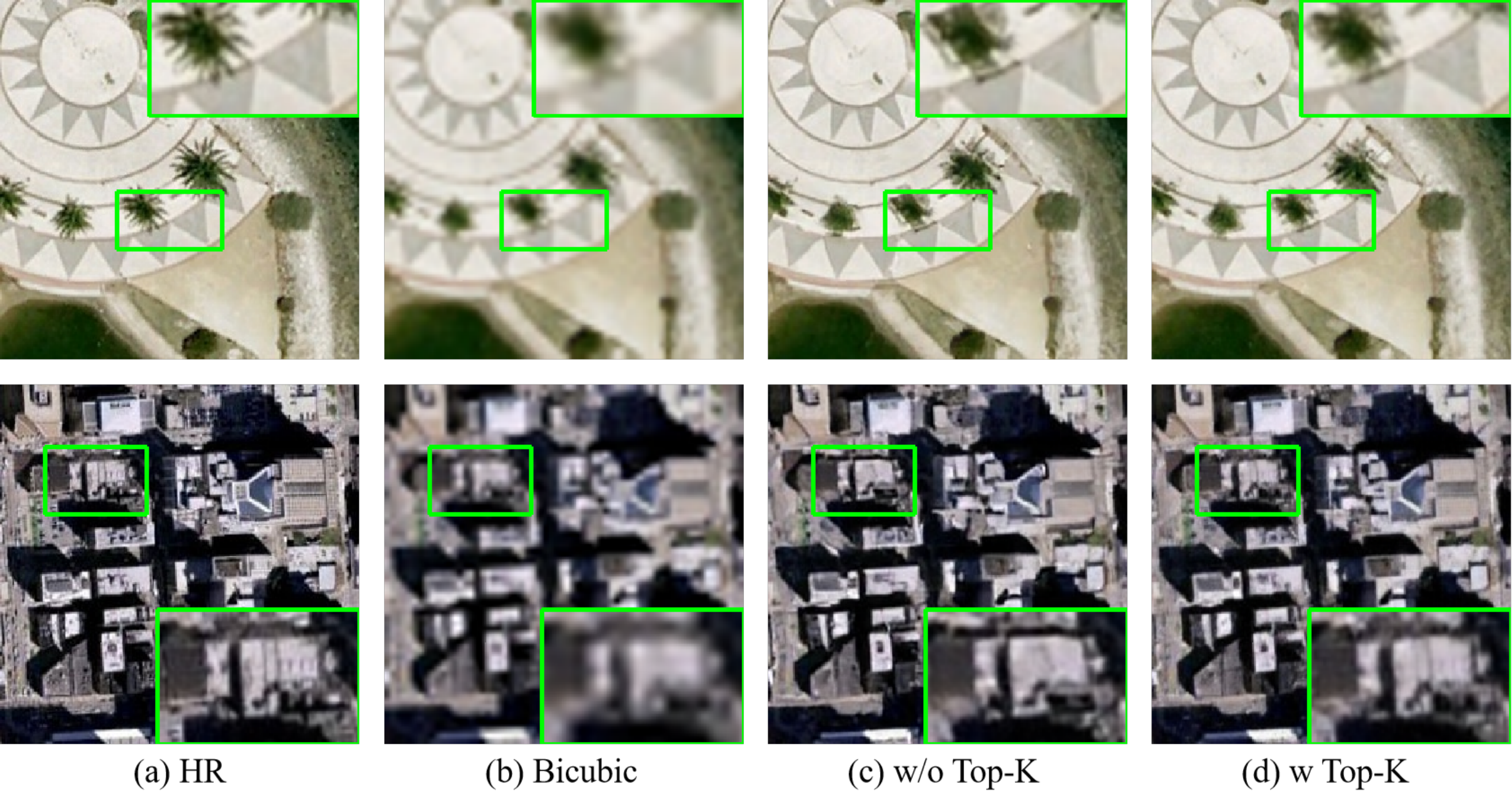} 
    \caption{Ablation visualization comparison of Top-K selection mechanism.}
    \label{figwotopk}
\end{figure}

To investigate the trade-off relationship of the Top-K selection mechanism between noise suppression and detail preservation, this study designed comparative experiments with and without the Top-K mechanism. From the SR results, the SR image without the Top-K mechanism exhibited more artifact information, failing to adequately restore building edges and authentic tree textures (Fig \ref{figwotopk}(c)). In contrast, the SR image with the Top-K mechanism achieved clear reconstruction of buildings and trees (Fig \ref{figwotopk}(d)).

\subsubsection{Validation of the Necessity of Dynamic Top-K Selection}
\begin{figure}[t!]  
    \centering  
    \includegraphics[width=0.50\textwidth]{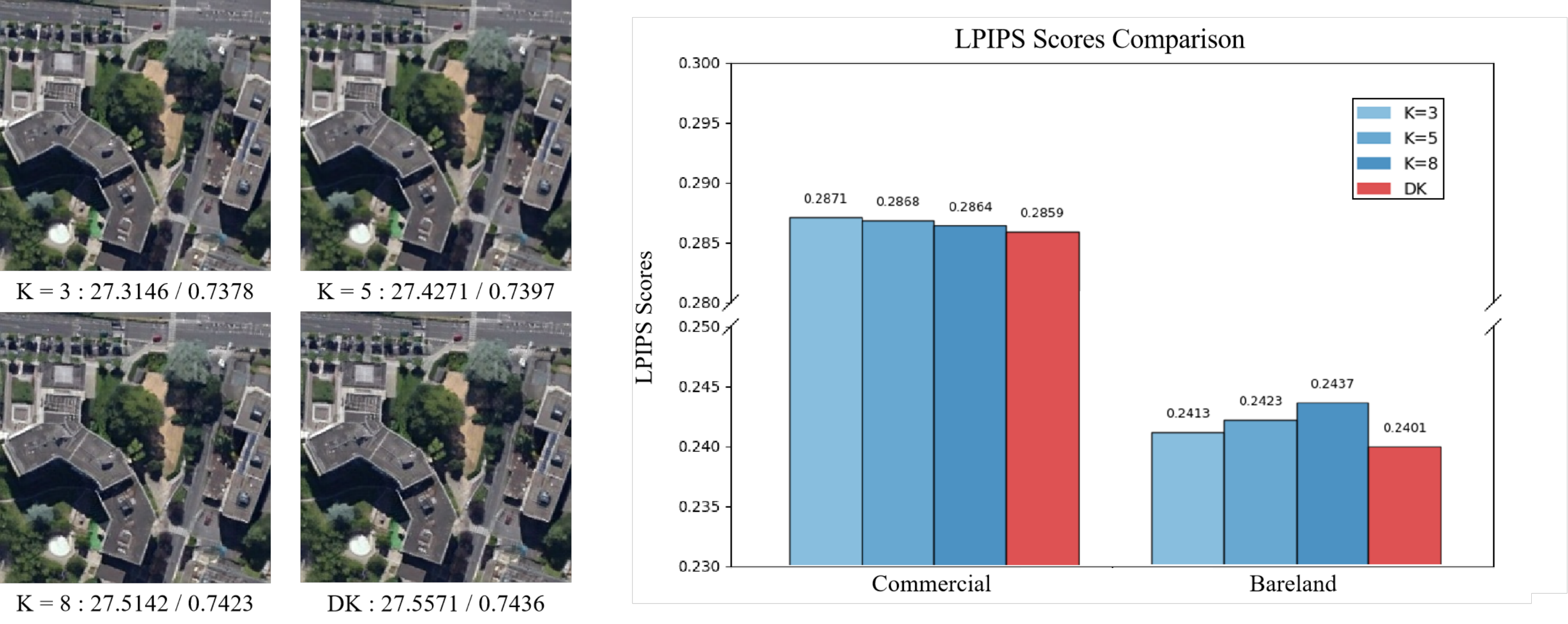} 
    \caption{Performance Comparison between Fixed K-value and Dynamic K-value in Complex Scenarios versus Uniform Scenarios. Subscript numbers denote the PSNR/SSIM score for this example image.}  
    \label{figtongji}  
\end{figure}

To investigate the effectiveness of the dynamically adjusted Top-K feature selection mechanism, we compared the performance of fixed-K and dynamic-K strategies in complex scenarios (commercial) and homogeneous scenarios (bareland). Experimental results demonstrate that in commercial scenarios with rich structural features, higher K-values effectively integrate more detailed information from reference images, significantly improving the LPIPS metrics of reconstructed images. Conversely, in homogeneous scenarios like bareland with uniform textures, higher K-values tend to introduce additional noise, leading to degraded LPIPS performance. Figure \ref{figtongji} further validates the scenario adaptability of our method: reconstruction quality in commercial areas matches the performance of fixed K=8 strategy, while in bareland regions it approaches the optimal fixed K=3 strategy. This complexity-aware dynamic adjustment mechanism effectively addresses the performance fluctuations of fixed-K strategies in cross-scenario applications, demonstrating robust adaptability for reference-based image SR tasks.

\subsection{Semantic Relevance Analysis of Reference Images}

To investigate the impact of reference image semantic relevance on the SR performance of TTRD3, this experiment covers three typical scenarios: strongly relevant reference (HR images from the same scene), weakly relevant reference (HR images from different scenes), and noise reference (Gaussian noise). The quantitative analysis results in Table \ref{tab5} show that when using strongly relevant reference images, the model achieves optimal performance, outperforming the metrics of weakly relevant reference images (PSNR: +0.3610 dB, SSIM: +0.0157, LPIPS: -0.0098, FID: -2.5074), which validates the effectiveness of the multi-scale texture transfer mechanism in extracting and fusing semantically matched features. 

Notably, even under weakly relevant reference conditions, the model maintains a performance advantage compared to the noise reference (PSNR: +0.0145 dB, SSIM: +0.0012, LPIPS: -0.0046, FID: -0.6413), indicating that cross-scene feature transfer possesses a certain degree of generalization capability. When the reference image degenerates into Gaussian noise, the model reverts to RDDM, and its performance metrics are highly consistent with those of the no-reference baseline model Model-1. This phenomenon confirms the strong robustness characteristics of TTRD3.

\begin{figure*}[t!]  
    \centering
    \includegraphics[width=1\textwidth]{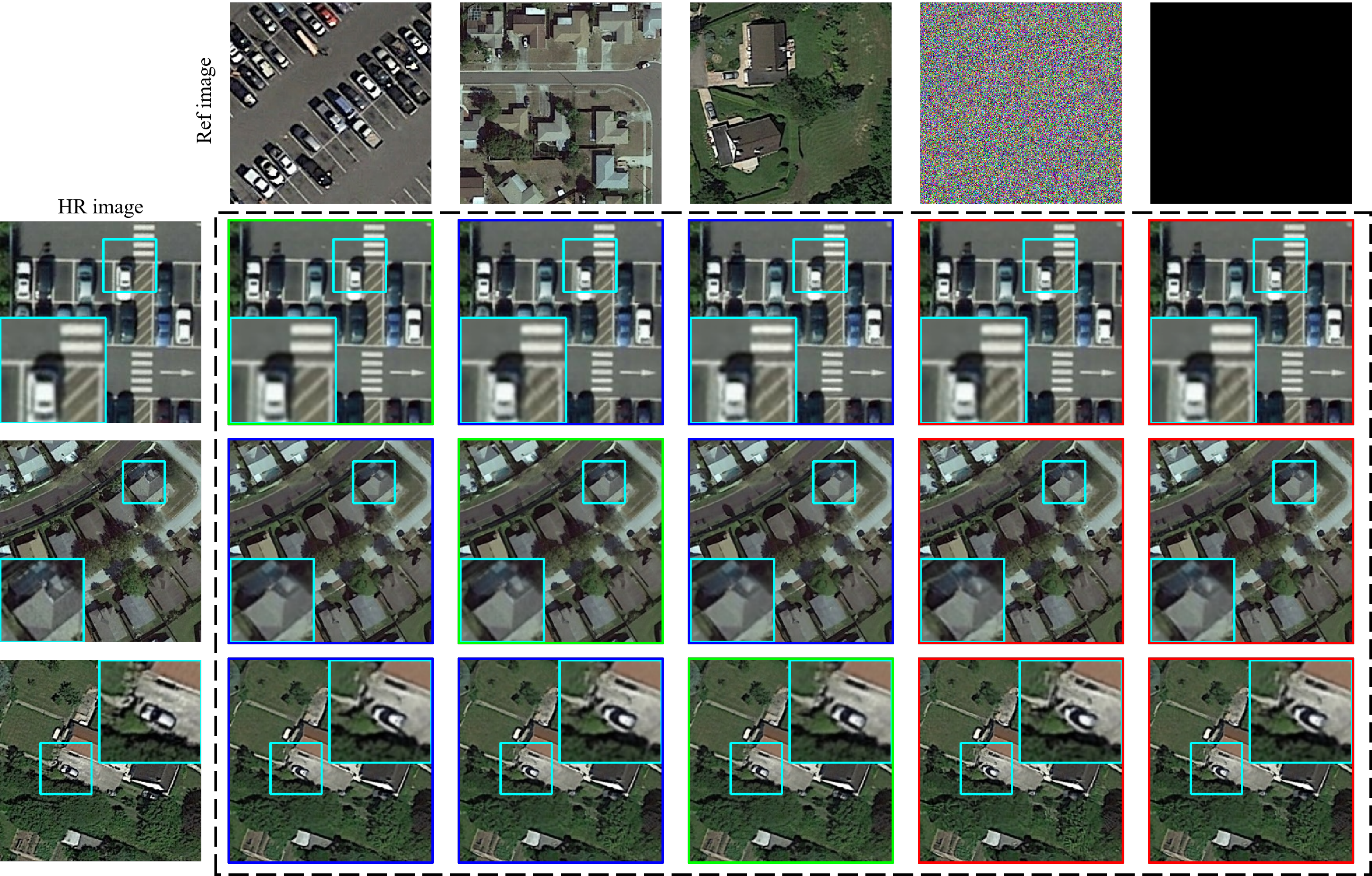}  
    \caption{Comparison of the impact of different reference images on SR results, where the SR images generated from strongly relevant reference images are marked with {\textcolor{green}{green}} boxes, the SR images generated from weakly relevant reference images are marked with {\textcolor{blue}{blue}} boxes, and the SR images generated from noise reference image and black reference image are marked with {\textcolor{red}{red}} boxes.}
    \label{figdiffref}
\end{figure*}

Figure \ref{figdiffref} demonstrates the impact of different levels of semantic relevance in reference images on SR results. When the reference image is strongly semantically relevant (same scene), the SR results exhibit optimal visual quality: car edges are sharp, zebra crossing textures are clear, and building structures are accurately aligned. When the reference image is weakly semantically relevant (different scene), the reconstruction quality slightly decreases, slightly blurred edges and misalignment. Even when the reference image is pure noise, the model maintains basic reconstruction capabilities, further validating its robustness.

\begin{table}[t!]
  \centering
  \caption{Comparison experiment results of reference images with different relevance levels in the AID test dataset}
     \resizebox{0.48\textwidth}{!}{
    \begin{tabular}{ccccc}
    \noalign{\hrule height 1.5pt}
    Reference Image & PSNR↑ & SSIM↑ & LPIPS↓ & FID↓ \\ \hline
    Relevant & 28.6798  & 0.7679  & 0.2531  & 33.6997 \bigstrut[t] \\
    Irrelevant & 28.3188  & 0.7522  & 0.2629  & 36.2071 \\
    Noise  & 28.3043  & 0.7510  & 0.2675  & 36.8484 \bigstrut[b] \\
    \noalign{\hrule height 1.5pt}
    \end{tabular}%
    }
  \label{tab5}%
\end{table}

\begin{figure}[t!]  
    \centering
    \includegraphics[width=0.49\textwidth]{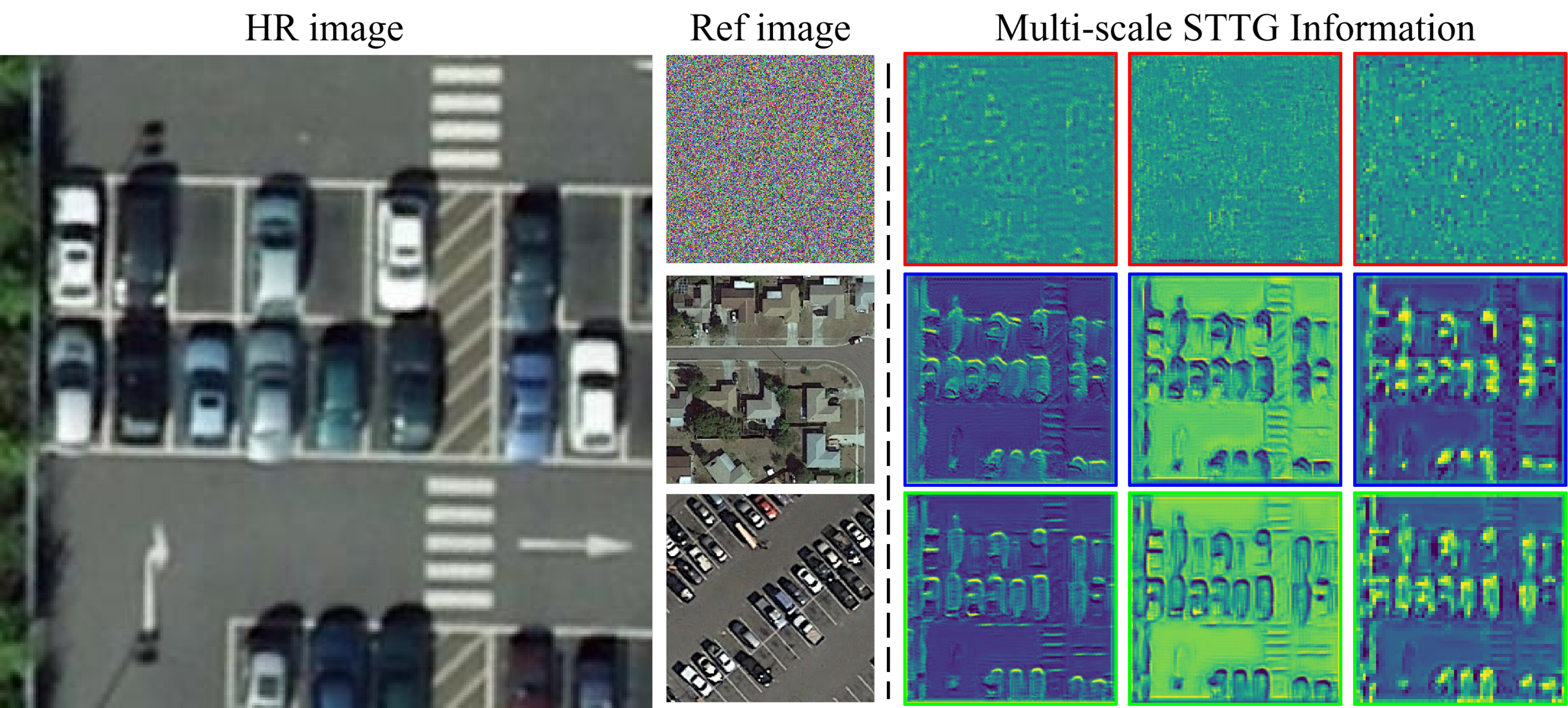}  
    \caption{Visualization of STTG output features for different reference images, where features from strongly relevant reference images are marked with {\textcolor{green}{green}} boxes, features from weakly relevant reference images are marked with {\textcolor{blue}{blue}} boxes, and features from noise reference images are marked with {\textcolor{red}{red}} boxes.}
    \label{figheat}
\end{figure}

To intuitively validate the guiding effect of reference images on texture transfer, this study conducted a visualization analysis of the multi-scale output features of the STTG module, with the results shown in Figure \ref{figheat}. By comparing the feature response maps of strongly relevant, weakly relevant, and noise reference images, significant differences can be observed: the feature maps of strongly relevant reference images exhibit clear structured texture patterns, with edge responses highly consistent with the target regions; although weakly relevant reference images can activate some features, their response patterns show obvious spatial dispersion; while the feature maps of noise reference images display uniformly distributed low-frequency noise, without revealing any effective texture structures. This visualization result confirms the critical role of HR reference images in the texture transfer process at the feature level, providing intuitive evidence for the feasibility of the multi-scale texture guidance mechanism.

\subsection{Comparative Analysis of Different Noise Intensities}
To investigate the impact of noise intensity on SR reconstruction performance during the diffusion process, we conduct comparative experiments on the AID test set with noise intensity parameters set to $\bar{\beta}_T=0.1$, $\bar{\beta}_T=0.5$ and $\bar{\beta}_T=1$.

Experimental results (Table \ref{tab4}) demonstrate that $\bar{\beta}_T=0.1$ achieves superior performance in perceptual quality metrics LPIPS and FID, while $\bar{\beta}_T=1$ significantly improves fidelity metrics PSNR and SSIM. When $\bar{\beta}_T=0.5$, the SR metrics are in a compromise between the noise with an intensity of 1 and the noise with an intensity of 0.1.

\begin{figure}[t!]  
    \centering  
    \includegraphics[width=0.49\textwidth]{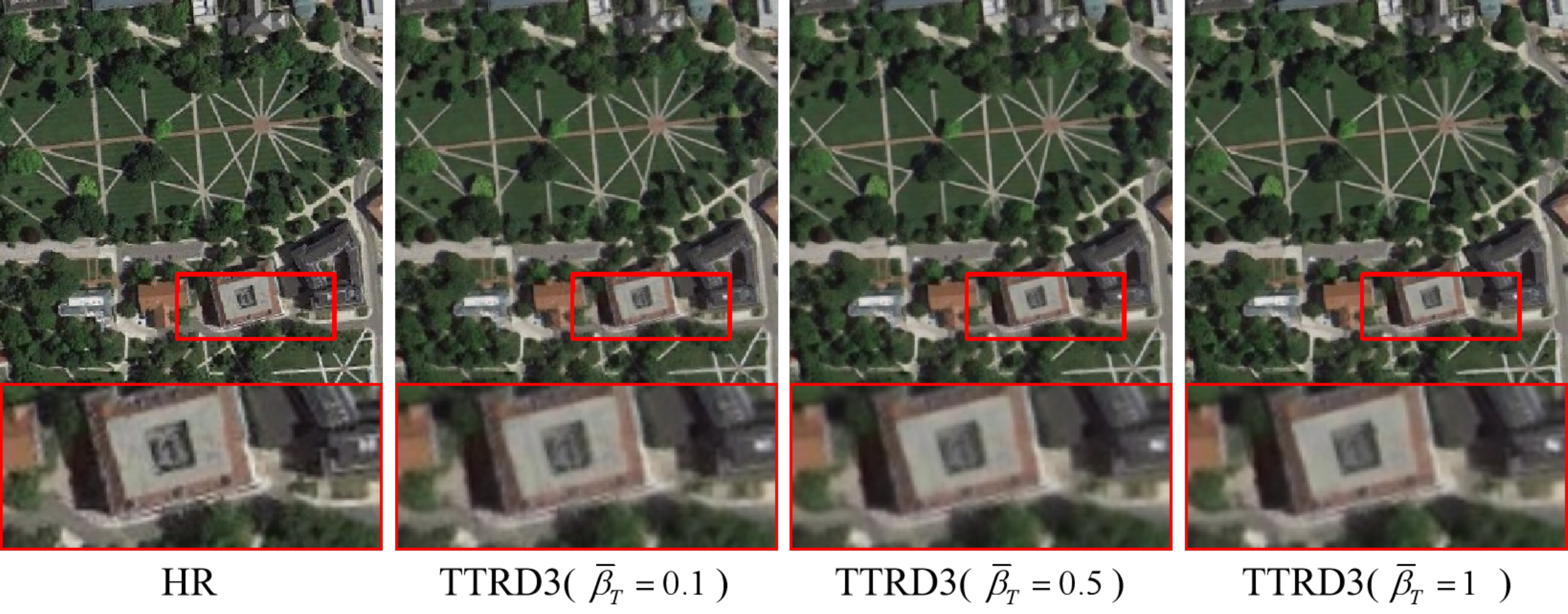}  
    \caption{Comparison of the Impact of Different Noise Intensities on the SR Results of TTRD3.}  
    \label{fig10} 
\end{figure}

Visual comparisons in Figure \ref{fig10} further reveal distinct reconstruction characteristics: images generated with high noise intensity exhibit a smoothing tendency, suitable for structure-sensitive scenarios, whereas low noise intensity restores richer texture details, producing results closer to real-world imagery. This inherent trade-off relationship between perceptual and fidelity metrics provides critical guidance for parameter optimization in practical applications.

\begin{table}[t!]
  \centering
  \caption{The comparative experimental results of noise intensity $\bar{\beta}_T=0.1$, $\bar{\beta}_T=0.5$ and $\bar{\beta}_T=1$ in TTRD3 on the AID test set.}
    \resizebox{0.48\textwidth}{!}{
    \begin{tabular}{>{\centering\arraybackslash}p{5em}
                     >{\centering\arraybackslash}p{4em}
                     >{\centering\arraybackslash}p{4em}
                     >{\centering\arraybackslash}p{4em}
                     >{\centering\arraybackslash}p{4em}}
    \noalign{\hrule height 1.5pt}
    Methods & PSNR↑ & SSIM↑ & LPIPS↓ & FID↓ \\ \hline
    $\bar{\beta}_T=0.1$ & 28.6798 & 0.7679 & 0.2531 & 33.6997 \bigstrut[t] \\
    $\bar{\beta}_T=0.5$ & 28.8628 & 0.7786 & 0.2613 & 36.5713 \\
    $\bar{\beta}_T=1$ & 29.0465 & 0.7819 & 0.2723 & 38.2344 \bigstrut[b] \\
    \noalign{\hrule height 1.5pt}
    \end{tabular}%
    }
  \label{tab4}%
\end{table}

\subsection{Analysis of the Residual Denoising U-Net Structure}

\begin{figure*}[t!]  
    \centering  
    \includegraphics[width=1\textwidth]{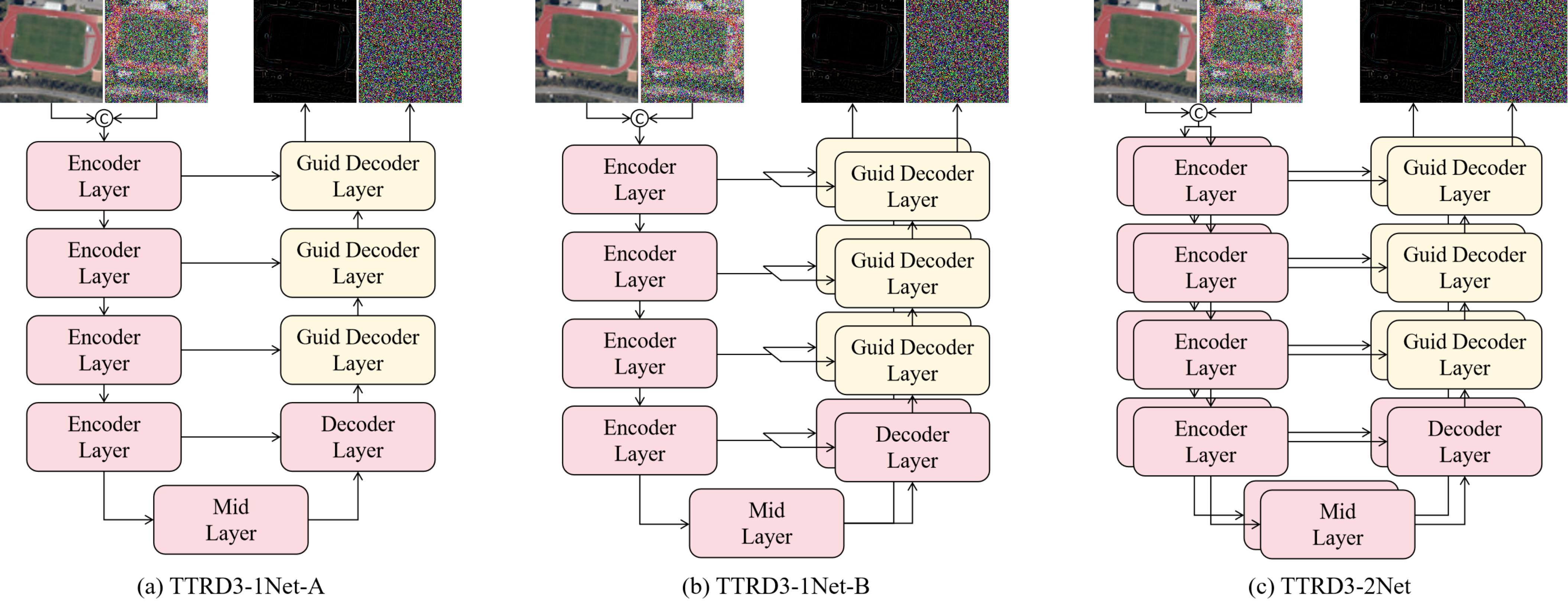}  
    \caption{Schematic of Three Residual Denoising Prediction Network Structures.}  
    \label{figUnet} 
\end{figure*}

In this study, an in-depth exploration of the U-Net architecture within the RDDM framework was conducted. Since RDDM generates SR images by jointly predicting residual information and noise information, its core prediction network can adopt either a single U-Net or dual U-Net architecture. For the single U-Net implementation, two distinct approaches were validated: the first is a fully shared architecture (TTRD3-1Net-A), where residual and noise predictions share the complete encoder-decoder path, with the network output layer expanded to six channels (the first three channels predict residuals, and the latter three predict noise). The second approach is an encoder-shared architecture (TTRD3-1Net-B), which uses a shared encoder for feature extraction but employs independent decoders to separately predict residuals and noise. As a comparative baseline, the dual U-Net independent architecture (TTRD3-2Net) utilizes completely separate networks for residual and noise prediction, as illustrated in Figure \ref{figUnet}.

\begin{table}[t!]
  \centering
  \caption{Comparison of Metrics for Three Residual Denoising Prediction Networks on the AID test set}
    \resizebox{0.48\textwidth}{!}{
    \begin{tabular}{ >{\centering\arraybackslash}p{6.5em}
                     >{\centering\arraybackslash}p{2.5em}
                     >{\centering\arraybackslash}p{2.5em}
                     >{\centering\arraybackslash}p{2.5em}
                     >{\centering\arraybackslash}p{2.5em}
                     >{\centering\arraybackslash}p{2.5em}
                     >{\centering\arraybackslash}p{5.9em}}
    \noalign{\hrule height 1.5pt}
    Methods & PSNR↑ & SSIM↑ & LPIPS↓ & FID↓ & Params & Inference time \bigstrut\\
    \hline
    TTRD3-1Net-A & 28.6946 & 0.7698 & 0.2559 & 37.5526 & 54.98M  & 0.2558s \bigstrut[t]\\
    TTRD3-1Net-B & 28.7538 & 0.7719 & 0.2542 & 36.3965 & 80.48M  & 0.3932s \\
    TTRD3-2Net & 28.6798  & 0.7679  & 0.2531  & 33.6997  & 110.11M & 0.5365s \bigstrut[b]\\
    \noalign{\hrule height 1.5pt}
    \end{tabular}%
    }
  \label{tab:unet}%
\end{table}%

Comparative experiments conducted on an NVIDIA RTX 4090 GPU (detailed in Table \ref{tab:unet}) revealed that the single U-Net approach significantly reduces computational costs at the expense of residual-noise decoupling. TTRD3-1Net-A and TTRD3-1Net-B reduced parameter counts by 50.07\% and 26.89\%, respectively, compared to TTRD3-2Net, while inference times per image decreased by 52.32\% and 26.93\%. However, parameter compression and feature coupling led to a decline in SR metrics: TTRD3-1Net-A exhibited the most significant performance degradation due to fully shared encoder-decoder layers. TTRD3-1Net-B partially preserved decoupling through independent decoders, achieving superior PSNR and SSIM metrics compared to the dual-network architecture. Although TTRD3-2Net achieved optimal LPIPS and FID scores, it required higher training costs and inference latency. Further analysis indicated that TTRD3-1Net-B achieves a relative balance between efficiency and performance, making it suitable for deployment in computationally constrained scenarios, while TTRD3-2Net is better suited for applications demanding rigorous reconstruction quality. These experimental results provide practical guidance for model selection based on hardware conditions and task objectives.

\begin{figure}[t!]  
    \centering  
    \includegraphics[width=0.5\textwidth]{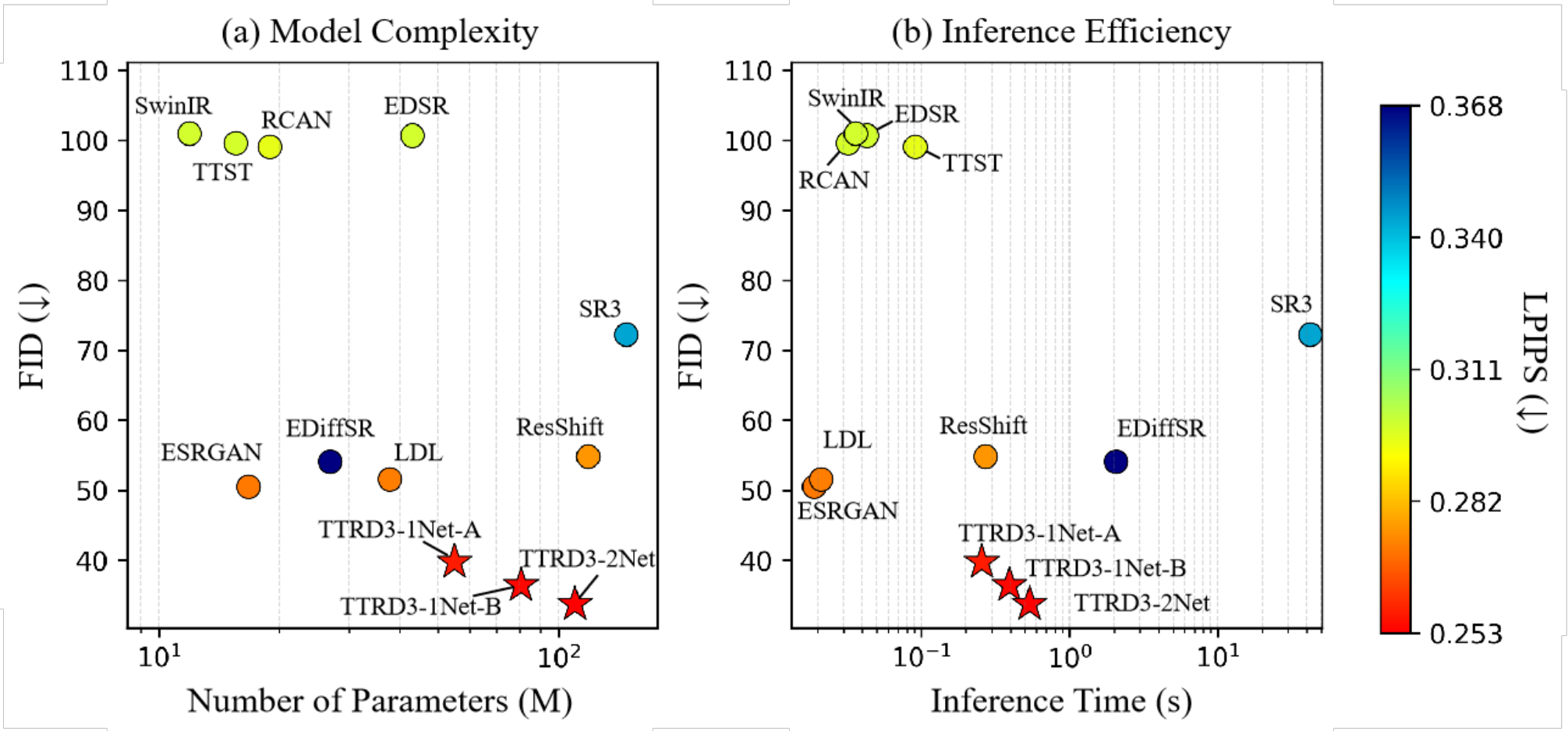}  
    \caption{In SR model efficiency evaluation, our TTRD3 method achieves a favorable trade-off between the number of parameters, inference time, and performance.} 
    \label{figcanshu} 
\end{figure}

\subsection{Estimation of Model Efficiency}

To validate the inference efficiency of TTRD3, this study evaluates the model from two perspectives: parameter count (unit in M) and per-image inference time (unit in s). All tests were conducted under a unified NVIDIA RTX 4090 GPU environment to ensure fairness, with results visualized in Figure \ref{figcanshu}. The x-axis represents parameter count and inference time, the y-axis denotes FID scores, and the color mapping of each method corresponds to LPIPS-based normalized scores (red indicates optimal performance).

Experimental results demonstrate that among DM-based methods, TTRD3-2Net achieves the best FID score (33.6997) with only 10 sampling steps, where its residual-noise dual prediction mechanism significantly reduces iterative requirements through feature complementarity. Although TTRD3-2Net’s parameter count (110.11M) is comparable to other DMs, the dual-diffusion synergy enables superior performance with fewer steps, achieving a per-image inference time of 0.537s. Compared to the pure residual prediction method ResShift (0.2713s inference time, FID 54.8146), TTRD3-2Net improves the FID metric by 38.5\%. When adopting the lightweight variant TTRD3-1Net-A, its inference time (0.2573s) outperforms ResShift, while the FID score (37.5526) also improves by 31.49\%. Notably, while current DMs frameworks lag behind single-step inference methods (such as CNNs, Transformers, and GANs) in computational efficiency due to iterative generation, they exhibit significant advantages in generative capability and visual quality.

\subsection{Comparative Experiment on Timesteps}

To investigate the impact of sampling steps on the performance of TTRD3, ablation experiments were conducted on TTRD3-1Net-A, TTRD3-1Net-B and TTRD3-2Net, testing step counts of 5, 10, 15, and 20. Metrics and inference times were recorded on an NVIDIA RTX 4090 GPU (Table \ref{tab:step}). Quantitative analysis showed that as sampling steps increased, SSIM and PSNR slightly declined, while perceptual quality metrics (LPIPS and FID) improved significantly, with inference times increasing approximately linearly. This finding underscores the necessity of progressive denoising for high-frequency detail recovery and provides critical guidance for parameter configuration: 5-10 steps balance quality and efficiency in time-sensitive scenarios, while 15-20 steps optimize visual fidelity for precision-critical tasks.

\begin{table}[t!]
  \centering
  \caption{Comparison of Metric Performance under Different Time Steps on the AID Test Set}
  \resizebox{0.48\textwidth}{!}{
    \begin{tabular}{
    >{\centering\arraybackslash}p{6.5em}
    >{\centering\arraybackslash}p{3em}
    >{\centering\arraybackslash}p{5.9em}
    >{\centering\arraybackslash}p{2.5em}
    >{\centering\arraybackslash}p{2.5em}
    >{\centering\arraybackslash}p{2.5em}
    >{\centering\arraybackslash}p{2.5em}}
    \noalign{\hrule height 1.5pt}
    Methods & Timesteps & Inference time & \multicolumn{1}{c}{PSNR↑} & \multicolumn{1}{c}{SSIM↑} & \multicolumn{1}{c}{LPIPS↓} & \multicolumn{1}{c}{FID↓} \bigstrut\\
    \hline
        \multirow{4}[2]{*}{TTRD3-1Net-A} & 5     & 0.1141s  & 29.1881  & 0.7868  & 0.2610  & 43.3455  \bigstrut[t]\\
          & 10    & 0.2558s  & 28.6946 & 0.7698 & 0.2559 & 37.5526 \\
          & 15    & 0.3921s  & 28.4529  & 0.7611  & 0.2562  & 36.1040  \\
          & 20    & 0.5305s  & 28.3051  & 0.7557  & 0.2569  & 35.6182  \bigstrut[b]\\
    \hline
    \multirow{4}[2]{*}{TTRD3-1Net-B} & 5     & 0.1863s  & 29.2176  & 0.7881  & 0.2633  & 42.8082  \bigstrut[t]\\
          & 10    & 0.3932s  & 28.7538 & 0.7719 & 0.2542 & 36.3965 \\
          & 15    & 0.6058s  & 28.5528  & 0.7645  & 0.2527  & 34.7272  \\
          & 20    & 0.8137s  & 28.4171  & 0.7595  & 0.2530  & 34.1448  \bigstrut[b]\\
    \hline
    \multirow{4}[2]{*}{TTRD3-2Net} & 5     & 0.2529s  & 29.1759  & 0.7870  & 0.2599  & 41.0092 \bigstrut[t]\\
          & 10    & 0.5365s  & 28.6798  & 0.7679  & 0.2531  & 33.6997 \\
          & 15    & 0.8125s  & 28.4462  & 0.7610  & 0.2523  & 32.3046 \\
          & 20    & 1.0911s  & 28.3092  & 0.7557  & 0.2516  & 31.6729 \bigstrut[b]\\    \noalign{\hrule height 1.5pt}
    \end{tabular}%
    }
  \label{tab:step}%
\end{table}%
\section{Conclusions}\label{sec6}
This study proposes TTRD3 for RSISR. The core RDDM framework integrates deterministic residual diffusion and stochastic noise diffusion to enhance texture diversity while preserving structural fidelity. To address the scarcity of high-frequency priors, MFAB fuses multi-scale features extracted by convolutional kernels of different scales, while STTG adaptively transfers semantically aligned high-frequency textures from reference images through a soft-hard attention collaborative mechanism. Evaluations demonstrate that TTRD3 outperforms SOTA methods in both fidelity and perceptual quality. Three architectural variants (TTRD3-1Net-A/B and TTRD3-2Net) are designed to flexibly adapt to different hardware conditions and task requirements, achieving a balance between high-precision reconstruction and fast processing.

Current limitations of TTRD3 include the fixed bicubic degradation assumption, which fails to effectively model composite degradations in real RS scenarios. Future work will extend TTRD3 to blind SR by incorporating degradation-aware mechanisms and kernel estimation.

\bibliographystyle{IEEEtran} 
\bibliography{ref}



\end{document}